\documentclass[lettersize,journal]{IEEEtran}
\usepackage{amsmath,amsfonts}
\usepackage{listings}
\usepackage{algorithmic}
\usepackage{bm}
\usepackage[ruled, vlined, linesnumbered]{algorithm2e}
\usepackage{array}
\usepackage{textcomp}
\usepackage{stfloats}
\usepackage{url}
\usepackage{verbatim}
\usepackage{graphicx}
\usepackage{cite}
\usepackage{multirow}
\usepackage{makecell}
\usepackage{booktabs}
\usepackage{color}
\usepackage{subcaption}
\usepackage{pifont}
\usepackage[flushleft]{threeparttable}  
\usepackage[table,dvipsnames]{xcolor}
\usepackage{wrapfig} 
\usepackage{xcolor}
\usepackage{multicol}
\usepackage[frozencache,cachedir=minted-cache]{minted}
\usepackage{pifont}
\usepackage[normalem]{ulem}

\newcolumntype{M}[1]{>{\centering\arraybackslash}m{#1}}

\def\ourlib{\texttt{EvoX}}

\newif\iffinal
\iffinal
  \newcommand{\zhuozhao}[1]{}
\else
  \newcommand{\zhuozhao}[1]{{\textcolor{blue}{ ZZ: #1 }}}
\fi

\lstdefinestyle{pythonstyle}{
  columns=fullflexible,
  breaklines=true,
  captionpos=b,
  xleftmargin=1em,
  backgroundcolor=\color{white},
  commentstyle=\color{green},
  keywordstyle=\color{magenta},
  numberstyle=\tiny\color{gray},
  stringstyle=\color{purple},
  basicstyle=\ttfamily\footnotesize,
  breakatwhitespace=false,
  breaklines=true,
  keepspaces=true,
  numbers=left,
  numbersep=5pt,
  showspaces=false,
  showstringspaces=false,
  showtabs=false,
  tabsize=2
}

\begin{document}

\title{EvoX: A Distributed  GPU-accelerated Framework for Scalable Evolutionary Computation}

\author{
        Beichen Huang,
        Ran Cheng, 
        Zhuozhao Li,
        Yaochu Jin,~\IEEEmembership{Fellow,~IEEE},
        and Kay Chen Tan,~\IEEEmembership{Fellow,~IEEE}
    \thanks{Beichen Huang was with the Department of Computer Science and Engineering, Southern University of Science and Technology, Shenzhen 518055, China. He is now with the Department of Computing, The Hong Kong Polytechnic University, Hong Kong SAR. E-mail: bill.huang2001@gmail.com.}
    \thanks{Ran Cheng and Zhuozhao Li are with the Department of Computer Science and Engineering, Southern University of Science and Technology, Shenzhen 518055, China. E-mails: ranchengcn@gmail.com, lizz@sustech.edu.cn. (\emph{Corresponding author: Ran Cheng}).}
    \thanks{Yaochu Jin is with the School of Engineering, Westlake University, Hangzhou 310030, China. Email: jinyaochu@westlake.edu.cn.}
    \thanks{Kay Chen Tan is with the Department of Computing, The Hong Kong Polytechnic University, Hong Kong SAR. E-mail: kctan@polyu.edu.hk.}
}

  \markboth{}
 {Shell \MakeLowercase{\textit{et al.}}}


\maketitle

\begin{abstract}
Inspired by natural evolutionary processes, Evolutionary Computation (EC) has established itself as a cornerstone of Artificial Intelligence. 
Recently, with the surge in data-intensive applications and large-scale complex systems, the demand for scalable EC solutions has grown significantly. 
However, most existing EC infrastructures fall short of catering to the heightened demands of large-scale problem solving. 
While the advent of some pioneering GPU-accelerated EC libraries is a step forward, they also grapple with some limitations, particularly in terms of flexibility and architectural robustness. 
In response, we introduce \ourlib{}: a computing framework tailored for automated, distributed, and heterogeneous execution of EC algorithms. 
At the core of \ourlib{} lies a unique programming model to streamline the development of parallelizable EC algorithms, complemented by a computation model specifically 
optimized for distributed GPU acceleration.
Building upon this foundation, we have crafted an extensive library comprising a wide spectrum of 50+ EC algorithms for both single- and multi-objective optimization.
Furthermore, the library offers comprehensive support for a diverse set of benchmark problems, ranging from dozens of numerical test functions to hundreds of reinforcement learning tasks.
Through extensive experiments across a range of problem scenarios and hardware configurations, \ourlib{} demonstrates robust system and model performances.
\ourlib{} is open-source and accessible at: \url{https://github.com/EMI-Group/EvoX}.
\end{abstract}


\section{Introduction}

\IEEEPARstart{I}{nspired} by the process of natural evolution, Evolutionary Computation (EC) has established its significance as a distinctive discipline within the expansive realm of Artificial Intelligence (AI)~\cite{back1997handbook}. 
EC's pivotal role is underscored by its inherent attributes: adaptability, resilience, and aptitude, which are pivotal for complex problem solving \cite{pena2000evolutionary, parmee2012evolutionary, malik2021metaheuristic}.   
Furthermore, EC plays a pivotal role in the quest for Artificial General Intelligence, especially within the context of neuroevolution~\cite{stanley2019designing}.

In the contemporary era, marked by large volumes of data and complex systems across diverse domains, the emphasis on \emph{scalability} in EC has gained paramount importance~\cite{omidvar2021review, liu2023survey}. 
As problem complexity and dimensionality increase, especially in fields like deep learning, there is an increasing demand for EC infrastructures to accommodate larger population sizes and higher-dimensional problems~\cite{liu2021survey, zhan2022evolutionary}.
Looking ahead, scalable EC will not only facilitate tackling larger problems but also pave the way towards emulating the complexity and adaptability inherent in biological systems~\cite{miikkulainen2021biological}. 
However, most existing EC infrastructures, including algorithm designs and computing frameworks, were merely tailored for smaller scales.
Thus, addressing the challenge of scalability is a critical research frontier in the ongoing development of EC.

Inherently, EC algorithms are well-suited to parallel computation due to their usage of a population of candidate solutions, each capable of independent evaluation. 
Consequently, they stand to benefit substantially from the parallel computation capabilities of hardware accelerators. 
Traditional CPU-based parallelization remains the prevailing approach, exemplified by DEAP~\cite{DEAP_JMLR2012}, PyGAD~\cite{gad2021pygad}, Pymoo~\cite{pymoo}, and Pagmo~\cite{Biscani2020}.
Until very recently, some pioneering advancements have led to the emergence of GPU-accelerated EC libraries such as EvoJAX~\cite{tang2022evojax}, evosax~\cite{lange2022evosax}, and EvoTorch~\cite{toklu2023evotorch}. 
Particularly, EvoJAX and evosax are based on JAX~\cite{jax2018github}, while EvoTorch is built upon PyTorch~\cite{pytorch} and Ray~\cite{moritz_ray_2018}.

Nonetheless, the pace of integrating hardware accelerators into EC infrastructures has been substantially slower compared to the strides made in the deep learning sector. 
A primary reason for this discrepancy is the lack of a universally endorsed and scalable computing framework.
Existing EC libraries, notably the recent GPU-accelerated ones, offer a plethora of useful tools and features. 
However, they are not without limitations. 
EvoJAX and evosax, in spite of harnessing GPU acceleration features, predominantly cater to evolution strategies with a focus on single-objective optimization, thus limiting their broader applicability. 
EvoTorch stands out with its adeptness in auto-parallelizing EC algorithms across multiple GPUs. 
However, its reliance on PyTorch, which is primarily tailored for deep learning rather than scientific computing, could impede its computational efficiency. 
Additionally, the parallelism of EvoTorch-based solutions is tightly coupled with the operators supported by PyTorch, potentially limiting the flexibility.
Most importantly, while EvoJAX, evosax, and EvoTorch are commendable as algorithm libraries, their architectural designs at the framework level are very limited. 
For example, they lack unified programming and computational models to enable simple implementation of general EC algorithms for seamless execution in distributed environments.

To address these limitations, we introduce \ourlib{}, a framework that facilitates automated, distributed, and heterogeneous execution of general EC algorithms. 
\ourlib{} implements a straightforward functional programming model, enabling users to declare the logical flow of an EC algorithm via a unified interface. 
Moreover, \ourlib{} adopts a hierarchical state management strategy that automatically distributes the tasks to arbitrary heterogeneous resources using a distributed execution engine. 
In summary, the main contributions are:
\begin{itemize}
    \item 
    We have designed and implemented \ourlib{}, a scalable and efficient framework that enables the execution of general EC algorithms across distributed heterogeneous systems.
    \item 
    We have proposed a straightforward functional programming model within \ourlib{} to streamline the development process of general EC algorithms for parallelization. 
    This model allows users to easily declare the logical flow of an EC algorithm, which reduces the complexity typically associated with the development process.
    \item 
    We have unified the main data stream and functional components into a flexible workflow. 
    This unification is achieved through a hierarchical state management module, which supports high-performance executions of general EC algorithms.
    \item 
    Leveraging the \ourlib{} framework, we have crafted a library that encompasses a wide spectrum of 50+ EC algorithms for both single- and multi-objective optimization. 
    Furthermore, the library has featured intuitive interfaces to diverse benchmark environments, comprehensively supporting hundreds of instances ranging from numerical optimization functions to reinforcement learning tasks.
\end{itemize}

The remainder of this paper is organized as follows.
Section~\ref{sec:relatedworks} presents some related work.
Section~\ref{sec:motivation} illustrates the motivation and requirements.
Section~\ref{sec:model} details the programming and computation models.
Section~\ref{sec:architecture} and Section~\ref{sec:implementation} elaborate on the architecture and implementation of \ourlib{} respectively.
Section~\ref{sec:experiments} conducts the experiments to assess the performance \ourlib{}, in comparison with EvoTorch.
Finally, Section~\ref{sec:conclusion} concludes the paper and discusses future work.

\section{Related Work}\label{sec:relatedworks}


\subsection{EC libraries in Python}

\textbf{DEAP}~\cite{DEAP_JMLR2012} stands as a comprehensive framework for EC algorithms in Python. 
With a rich history and a plethora of features, it caters to a broad spectrum of EC algorithms, encompassing both single- and multi-objective variants. 
Its collection of built-in benchmark problems facilitates the comprehensive evaluations of EC algorithms.

\textbf{PyGAD}~\cite{gad2021pygad} is a specialized platform for devising genetic algorithms in Python. 
It empowers users with a variety of crossover, mutation, and parent selection operators. 
Notably, it is particularly tailored for machine learning tasks, offering specialized tools and features for neural network training, cementing its role in integrating EC with machine learning pursuits.

\textbf{Pymoo}~\cite{pymoo} focuses primarily on multi-objective optimization problems, providing a robust platform for EC in Python. 
Its extensive support for diverse benchmark problems and state-of-the-art multi-objective EC algorithms, combined with visualization tools, highlights its commitment to the EC domain.

\textbf{Pygmo}~\cite{Biscani2020}, distinguished by its emphasis on massively parallel optimization, adopts the generalized island model for coarse-grained parallelization. 
It offers a vast array of algorithms and benchmark problems, facilitating the efficient deployment of parallelized EC algorithms.
Its batch fitness evaluation feature further enhances its utility.

\textbf{EvoJAX}~\cite{tang2022evojax} pushes the boundaries of scalable, hardware-accelerated neuroevolution. 
Leveraging the capabilities of the JAX framework, it seamlessly integrates evolution strategies with neural networks, ensuring efficient GPU parallelism. 
Building upon JAX, it provides a NumPy-like environment with just-in-time (JIT) compilation.

\textbf{evosax}~\cite{lange2022evosax}, a recent addition, positions itself as a dedicated library for GPU-accelerated ES. 
Taking cues from EvoJAX and deeply integrated with the JAX infrastructure, it presents a curated suite of ES algorithms, each of which is optimized for GPU performance.

\textbf{EvoTorch}~\cite{toklu2023evotorch} is another emerging library placing an emphasis on addressing scalability challenges within EC by harnessing the power of GPU acceleration. 
By seamlessly integrating with PyTorch, it not only gains the inherent advantages of this popular deep learning framework but also naturally taps into the vast resources of the Python community.

\subsection{JAX}
JAX~\cite{jax2018github} has rapidly ascended the ranks to become a leading computational infrastructure. 
While it offers a NumPy-style API, JAX has been optimized for GPU-accelerated numerical operations. 
JAX's unique selling point is its adaptability to contemporary computational challenges.
Its just-in-time (JIT) compilation transforms user-defined Python functions into high-performance machine code adaptable to diverse hardware platforms. 

One of JAX's notable features is its ability to fuse operators, thereby amalgamating multiple smaller tensor operations into singular and efficient tasks, thus reducing memory overhead. 
Its integrated autograd system enables automatic gradient computations, an essential feature for gradient-based optimization. 
Adhering to functional programming paradigms, JAX ensures computations are pure and side-effect-free, resulting in predictable and debug-friendly code. 
Such principles align well with EC algorithms, which are stateless by nature and suitable for a functional setting.
\section{Motivation and Requirements\label{sec:motivation}}

\begin{figure}
    \centering
    \includegraphics[width=\linewidth]{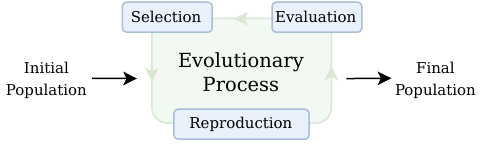}
    \caption{
    The typical process of an EC algorithm.
    Starting with an initial population, the EC algorithm engages in problem-solving through an iterative evolutionary process.
    Specifically, the main loop of this process evolves the population via three primary components: reproduction, evaluation, and selection.
    Ultimately, the final population is output as the solution set to the problem at hand.
    }
    \label{fig:ec-process}
\end{figure}

As shown in Fig.~\ref{fig:ec-process}, the typical process of an EC algorithm involves evolving a \emph{population} of candidate solutions to a specific problem.
In each iteration of the main loop, the current population first undergoes a \emph{reproduction} phase to generate new candidate solutions.
Next, the \emph{evaluation} phase assesses the effectiveness, a.k.a. \emph{fitness}, of each candidate solution in solving the problem at hand.
Following evaluation, a \emph{selection} process is carried out, favoring candidate solutions with superior fitness for inclusion in the subsequent generation, in line with the principle of \emph{survival of the fittest}.
This process repeats until either a satisfactory solution or set of solutions is found, or a predetermined number of iterations are completed.

From the perspective of a computing framework, the \emph{population} can be viewed as the primary data flow, while the \emph{reproduction}, \emph{evaluation}, and \emph{selection} are three functional components. 
Therefore, a computing framework designed for scalable EC must efficiently support these workloads. 
We provide a brief description of these workloads below.
\begin{itemize}
\item \emph{Population}: 
This object, consisting of candidate solutions in various data structures, goes through each functional component of the entire evolutionary process. 
In some emerging applications of EC such as deep neuroevolution~\cite{stanley2019designing}, the population may require computationally expensive decoding for complex representations.
\item \emph{Reproduction}: 
This process, such as the crossover/mutation operators in a genetic algorithm, often consists of a set of heuristic strategies for generating new candidate solutions. 
Reproduction can be parallelized in most EC algorithms that adopt dimension-wise operations when generating each new candidate solution.
\item \emph{Evaluation}: 
This process can be intrinsically parallelized in a distributed setting as the evaluation of each candidate solution is independent. 
In some emerging applications of EC such as evolutionary reinforcement learning~\cite{bai2023evolutionary}, the computing task may require heterogeneous hardware (i.e., CPUs \& GPUs) for hybrid complex simulations and deep learning tasks.
\item \emph{Selection}: 
This process is often realized via a set of sorting/ranking strategies, which can be compute-intensive with respect to population size. 
Parallelized selection is particularly beneficial in EC paradigms involving multiple populations.
\end{itemize}


Unlike deep learning, which benefits from a streamlined end-to-end workflow, the various components of an EC workflow, including data flow processing and functional elements, play distinct yet interconnected roles. 
This uniqueness poses significant challenges when designing a general computing framework to support EC algorithms. 
Moreover, as the applications of EC continue to expand, a widening gap has emerged between EC and other major branches of AI.
Bridging this gap necessitates the development of a distributed computing framework that can efficiently handle EC workloads, ensuring scalability and compatibility with heterogeneous computing environments. 
Specifically, such a framework should meet the following requirements.

\begin{itemize}
\item \emph{Flexibility and Extensibility}: 
The framework should support easy implementation of a broad spectrum of EC algorithms and black-box optimization problems. 
To this end, it should provide flexibility in the representation of candidate solutions (e.g., strings, trees, neural networks), the combination of various reproduction and selection mechanisms, as well as diverse methods for evaluation. 
Users should be able to leverage the framework to integrate existing modules seamlessly, thereby constructing workflows that align with their specific research goals and problem contexts.

\item \emph{Parallelizable Programming}: 
The framework should adopt a programming model geared towards parallelization. 
While this model offers advantages such as improved code clarity, testability, and reusability, its primary strength lies in its innate ability to support parallel execution. 
Consequently, even EC algorithms developed without direct parallel considerations should intuitively harness this capability, thus allowing for efficient execution across multiple computing units concurrently.

\item \emph{Heterogeneous Execution}: 
The framework should enable task executions across various types of computing resources, including CPUs, GPUs, or even other specialized hardware. 
It should also support distributed computing across multiple nodes, which involves intelligent data sharding, task scheduling, and resource management based on the requirements of tasks and available hardware. 
It should transparently manage data transfer and synchronization between different nodes.
\end{itemize}

In response to these requirements, we design and implement \ourlib{}. 
In the following section, we will elaborate on the programming and computation models underpinning this framework.

\section{Programming and Computation Models\label{sec:model}} 

\subsection{Programming Model}

\begin{listing}
\begin{minted}[
    fontsize=\footnotesize,
    linenos,
    frame=lines,
    framesep=2mm,
    xleftmargin=6pt,
    autogobble,
    numbersep=2pt
]{python}
from evox import Algorithm, State

class EC(Algorithm):
  def __init__(self, reproduction, selection, ...):
    # initialize operators
    super().__init__()
    self.reproduce = reproduction
    self.select = selection
    ...

  def setup(self, key):
    # initialize data
    init_population = ...
    init_fitness = ...
    ...
    # initialize state
    return State(
      pop = init_population,
      fit = init_fitness,
      ...
      key=key
    )

  def ask(self, state):
    # generate offspring
    offspring = self.reproduce(state.pop, ...)
    # update state 
    state = state.update(
      # merge offspring into the population
      pop = merge(pop, offspring),
      ...
    )
    return offspring, state

  def tell(self, state, fitness):
    # select new population
    new_pop, new_fit = self.select(
      state.pop, merge(state.fit, fitness))
    # update state
    state = state.update(
      pop = new_pop,
      fit = new_fit,
      ...
    )
    return state
\end{minted}
\caption{An exemplar implementation of a vanilla EC algorithm within the \ourlib{} framework. This implementation comprises four sections: \texttt{\_\_init\_\_}, \texttt{setup}, \texttt{ask}, and \texttt{tell}. The \texttt{\_\_init\_\_} section is dedicated to initializing foundational operators, including reproduction and selection mechanisms. The \texttt{setup} segment focuses on data initialization, encompassing elements like the population data and fitness data. Within the \texttt{ask} section, the offspring are produced utilizing the reproduction operator, while the \texttt{tell} section facilitates population updates through the selection operator.}
\label{lst:algorithm_programming_model}
\end{listing}

To meet the requirement of \emph{Parallelizable Programming} (outlined in Section~\ref{sec:motivation}), we adopt the \emph{functional programming} paradigm as the core of our programming model.
This design is rooted in the paradigm's intrinsic benefits, including streamlined code, augmented reusability and testability, and a natural affinity for parallel and distributed computing. 
With this model, our aim is to provide an intuitive interface that facilitates the straightforward implementation of a wide spectrum of EC algorithms and their corresponding workflows.

An illustrative example of using the proposed programming model to implement a vanilla EC algorithm is presented in Listing~\ref{lst:algorithm_programming_model}.  
This implementation comprises four sections: \texttt{\_\_init\_\_}, \texttt{setup}, \texttt{ask} and \texttt{tell}. 
Within the \texttt{\_\_init\_\_} section, basic operators essential to the EC algorithm, encompassing processes like reproduction and selection, are defined. 
Given their invariant nature throughout the evolutionary trajectory, these operators are archived under the Python attribute \texttt{self}. 
The \texttt{setup} section is devoted to initializing essential data elements such as the population data and fitness data, as well as a key for random number generation. 
As these elements possess a mutable characteristic and are subject to modifications with each iteration, they are aptly demarcated as part of the \texttt{state}.
Subsequently, the \texttt{ask} section is tailored for the generation of offspring, realized through the designated reproduction operator (e.g., crossover and mutation). 
By contrast, the \texttt{tell} section orchestrates the assimilation of offspring into the prevailing population, culminating in an updated population based on the predefined selection operator.
Table~\ref{tab:api_summary} provides a concise overview of the primary modules and functions within the programming model, which are detailed as follows.

\begin{table}[]
    \centering
    \caption{Summary of core modules and functions in \ourlib{}'s programming model.}
    \begin{tabular}{l l l}
        \toprule
        Module & Function & Description \\
        \toprule
        \multirow{2}{4em}{Algorithm} & \texttt{ask} & To generate offspring population.\\
        \cmidrule(lr){2-3}
                                    & \texttt{tell} & To select new population.\\
        \midrule
        Problem & \texttt{evaluate} & To evaluate the fitness of a given population. \\
        \midrule
        \multirow{2}{4em}{Monitor} & \texttt{record\_pop} & To record the population data.\\
        \cmidrule(lr){2-3}
                                   & \texttt{record\_fit} & To record the fitness data.\\
        \midrule
        \multirow{2}{4em}{Workflow} & \texttt{init} & To initialize the workflow.\\
        \cmidrule(lr){2-3}
                                    & \texttt{step} & To execute one iteration of the workflow.\\
        \bottomrule
    \end{tabular}

    \label{tab:api_summary}
\end{table}

\textbf{Algorithm} is tailored for encapsulating EC algorithms. 
At its core is the \texttt{ask-and-tell} interface, which conceptualizes the EC algorithm as an agent in perpetual interaction with a problem at hand, involving \texttt{ask} and \texttt{tell} functions:
\begin{itemize}
    \item \texttt{ask}: it processes the \texttt{state} to yield new candidate solutions, updating the algorithmic \texttt{state} by merging these solutions with the existing population.
    \item \texttt{tell}: after ingesting the \texttt{state} and the fitness metrics of the new offspring, it selects candidate solutions for the next iteration, resulting in a refreshed algorithmic \texttt{state}.
\end{itemize}

\textbf{Problem} is tailored for modeling the problems to be solved. Central to this module is the \texttt{evaluate} function dedicated to determining the fitness values of candidate solutions within a population. 
Notably, the \textbf{Problem} module is constructed as a stateful procedure, such that it is capable of facilitating the management of intricate external environments. 
For instance, neuroevolution tasks reliant on external datasets can use the \textbf{Problem} module to manage the current batch via its state.

\textbf{Monitor} serves as an optional module for monitoring the data flow when running an EC algorithm. 
Utilizing callback functions, the \textbf{Monitor} module receives user-specified data by employing callback function, e.g., \texttt{record\_pop} and \texttt{record\_fit} for population data and fitness data respectively. 
Within the \textbf{Monitor} module, users can further process the data by statistical means or visualization tools.

\textbf{Workflow} plays a pivotal role in integrating various components to shape a cohesive and executable EC workflow. 
It begins by amalgamating diverse modules of \textbf{Algorithm}, \textbf{Problem}, and \textbf{Monitor} specified by users. 
Following this, it acts as the primary module, sequentially initiating every component via the \texttt{init} function.
This orchestration produces a global state, enveloping individual states as defined by each module's \texttt{setup} method.
Listing~\ref{lst:workflow_programming_model} showcases the usage of the workflow within \ourlib{}. 
After selecting an algorithm, a problem, and a monitor, users combine these instances within a workflow object.
Post-initialization via \texttt{init}, which activates every module under the workflow object recursively, users can fine-tune the workflow for advanced computational tasks, thus facilitating the efficient execution of the workflow across multiple nodes. 
The workflow's execution is then triggered using the \texttt{step} function.
Each call will take a single step in the iteration\footnote{Within the realm of EC algorithms, one iteration does not necessarily correspond to a single generation. Specifically, an iteration denotes a solitary \texttt{ask-evaluate-tell} loop, while some algorithms may require several iterations to complete a single generation.}.

\begin{listing}
\begin{minted}[
    fontsize=\footnotesize,
    linenos,
    frame=lines,
    framesep=2mm,
    xleftmargin=6pt,
    autogobble,
    numbersep=2pt
]{python}
from evox.workflow import StdWorkflow

# workflow creation
workflow = StdWorkflow(
  algorithm,
  problem,
  monitor
)

# workflow initialization
key = ... # A random number generator key
state = workflow.init(key)

# workflow analyzer (optional)
state = workflow.enable_multi_devices(state)
state = workflow.enable_distributed(state)


while ...: # not terminated
  # workflow runner
  state = workflow.step(state) # one iteration
\end{minted}
\caption{Illustrative usage of a workflow in \ourlib{}.
First, a workflow object is created.
Second, all components within a workflow are initialized, and a global state is returned.
Then the workflow can be configured with the help of a workflow analyzer.
Finally, the workflow is executed inside a loop in a distributed environment. 
}
\label{lst:workflow_programming_model}
\end{listing}

\subsection{Computation Model}\label{sec:model_computation}

\begin{figure}
    \centering
    \includegraphics[width=\columnwidth]{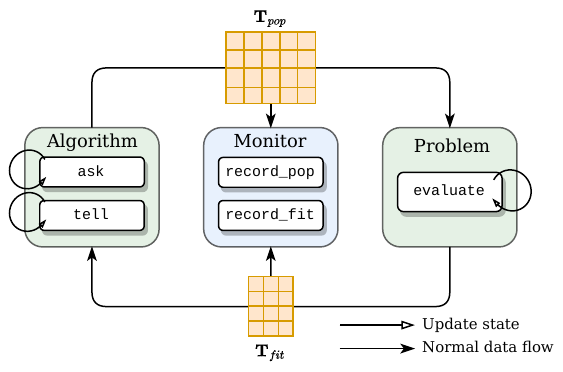}
    \caption{Illustration of the computation model adopted by \ourlib{}.
    There are three main modules: \textbf{Algorithm}, \textbf{Problem}, and \textbf{Monitor}.
    The iteration starts with \texttt{ask} function which generates $\textbf{T}_{pop}$ as the tensorized population.
    Then the population is sent to the \textbf{Problem} module for fitness evaluations via \texttt{evaluate} function and then generate  $\textbf{T}_{pop}$ as the tensorized fitness.
    Finally, the fitness is passed back to the \textbf{Algorithm} module through the \texttt{tell} function.
    Meanwhile, $\textbf{T}_{pop}$ and $\textbf{T}_{fit}$ can be optionally sent to the \textbf{Monitor} module for further processing, where \texttt{record\_pop} and \texttt{record\_fit} functions record $\textbf{T}_{pop}$ and $\textbf{T}_{fit}$ respectively.
    In addition to the normal data flow, each module can update its individual state at every function call.
    }
    \label{fig:computation_model}
\end{figure}

\ourlib{} employs a \emph{workflow} abstraction to perform computations and automatically invoke different instances (e.g., \textbf{Algorithm} and \textbf{Problem} modules) when they are ready to execute. 
In this subsection, we detail how \ourlib{} translates a user-written program (Listing~\ref{lst:workflow_programming_model}) into an automated workflow (Fig.~\ref{fig:computation_model}) for parallel execution internally.
At the core, an EC workflow is driven by executing an \texttt{ask-evaluate-tell} loop iteratively, which seamlessly integrates the \textbf{Algorithm}, \textbf{Problem}, and \textbf{Monitor} modules.

Initially, each iteration commences with the \texttt{ask} function in the \textbf{Algorithm} module to generate a population. 
To harness the capabilities of hardware acceleration, this population is encoded in a \emph{tensor} format, i.e., tightly packed multi-dimensional arrays in memory for data-level parallelism.
For simplicity, we denote the tensorized population data as $\textbf{T}_{pop} =  [\textbf{x}_1, \textbf{x}_2, ..., \textbf{x}_i, ...]$ hereafter, where $\textbf{x}_i$ denotes each encoded candidate solution.

Upon generating the population, the \texttt{evaluate} function in the \textbf{Problem} module will return the fitness data, i.e., the corresponding fitness values of the candidate solutions within the population. Similar to $\textbf{T}_{pop}$, the fitness data adopts a tensor representation to streamline subsequent processing.
For simplicity, we denote the tensorized fitness data as $\textbf{T}_{fit} = [\textbf{y}_1, \textbf{y}_2, ..., \textbf{y}_i, ...]$ hereafter, where $\textbf{y}_i$ denotes the fitness value of each candidate solution in the population.

Finally, the loop enters the \texttt{tell} function of the \textbf{Algorithm} module, which proceeds to select candidate solutions for generating new $\textbf{T}_{pop}$ according to $\textbf{T}_{fit}$. Given the tensorial nature of both $\textbf{T}_{pop}$ and $\textbf{T}_{fit}$, the \textbf{Algorithm} module can easily adapt to data-level parallelism.

Outside this primary loop, each module is responsible for two distinct data categories: immutable data and mutable data (i.e., the \texttt{state}). Immutable data, often initialized during object instantiation via Python's \texttt{\_\_init\_\_} method, encompass static hyperparameters pertinent to each module and might include external datasets for specific problems. These remain unchanged throughout the computational life-cycle.
In contrast, the \texttt{state} is initialized after the entire workflow has been assembled, starting with the topmost module and proceeding recursively through the dependency tree. 
During computation, as the global state is activated, state management ensures that each module receives its specific state data as the primary argument for its function. 
This segmented strategy enables each module to concentrate on updating its own state, while collaboratively participating in the comprehensive update of a unified global state.

One of the distinguishing attributes of \ourlib{} is its foundational alignment with the tenets of functional programming.
This approach accentuates a distinct separation between data and functions, thus ensuring that functions remain free from side effects. 
From this perspective, executing an EC algorithm can be perceived as orchestrating state updates via a coherent data flow, which can be formally articulated as follows.

Given $\theta$ as the hyperparameters, an EC algorithm $\mathcal{A}_{\theta}$ can be denoted as $\langle \theta, g^\text{ask}(\cdot), g^\text{tell}(\cdot) \rangle$, where  $g^\text{ask}(\cdot)$ and $g^\text{tell}(\cdot)$ correspond to the \texttt{ask} and \texttt{tell} functions, respectively.
When $\mathcal{A}_{\theta}$ is applied to a specific problem $\mathcal{P}_\mathcal{D}$ with $f^\text{evl}(\cdot)$ as its \texttt{evaluate} function, $\mathcal{D}$ comprises the parameters or datasets related to the problem. 
At each step, the data flow is articulated as:
\begin{align}
    \textbf{T}_{pop}, S &= g_{\theta}^\text{ask}(S), \\
    \textbf{T}_{fit}, S &= f^\text{evl}(S, \textbf{T}_{pop}), \\
    S&= g_{\theta}^\text{tell}(S, \textbf{T}_{fit}),
\end{align}
where $S$ denotes the $\texttt{state}$ that encapsulates the mutable data associated with $\mathcal{A}_{\theta}$ and $\mathcal{P}_\mathcal{D}$.
From this formulation, it is evidenced that the only variable undergoing updates at each iterative step is the state \( S \). 
Consequently, the operation of the entire EC algorithm can be envisioned as a sequence of state transitions, which is realized through the repeated execution of the \texttt{step} function within the \textbf{Workflow} module.

Beyond the primary computational processes of EC, the \textbf{Monitor} module also plays a pivotal role. 
Users are usually interested in the data generated during the running process of an EC algorithm, such as the best solution found so far, and the distribution of the population, among various other aspects. 
The \textbf{Monitor} module tailored this purpose in a \emph{pluggable} and \emph{asynchronous} manner. 
When the user plugs a monitor into the workflow, both $\textbf{T}_{pop}$ and $\textbf{T}_{fit}$ are asynchronously sent to the monitor. 
This allows the main EC computational process to progress uninterrupted, without waiting for the monitor to complete the data processing. 
Given that the monitor might involve subsequent slow disk I/O or plotting functions, this asynchronous design can significantly enhance the overall computational efficiency. 
Furthermore, since the monitor does not interfere with the primary computational process, the module even allows users to employ multiple monitors in parallel when needed.

To realize our envisioned workflow abstraction which coordinates these different modules, \ourlib{} employs a \emph{stateful computation} model for all primary modules. 
This model adopts the signature \texttt{(state, ...) -> (result, state)}, where each module operates based on its current state and other inputs, subsequently producing a result and an updated state. 
However, as we arrange these modules hierarchically, complexities arise: although it is naturally expected for a module to manage only its state, it also inherits the responsibility for the states of its nested sub-modules. 
To meet the requirement of such layered state management, \ourlib{} adopts a \emph{hierarchical state management} mechanism, ensuring the smooth crafting of stateful computations across intricate hierarchies. 
Central to this system is the \texttt{state} variable, which acts as a \emph{universal conduit} across computational modules. 
Rather than being a static entity, this variable dynamically adjusts to encapsulate the relevant state during each function's execution. 
This adaptive design ensures automated state transitions and efficient management across the comprehensive hierarchical framework. 
In the following section, we will delve into the hierarchical state management mechanism, as well as other main components that underpin the architecture of \ourlib{}.

\section{Architecture \label{sec:architecture}}

\begin{figure}
    \centering
    \includegraphics[width=\linewidth]{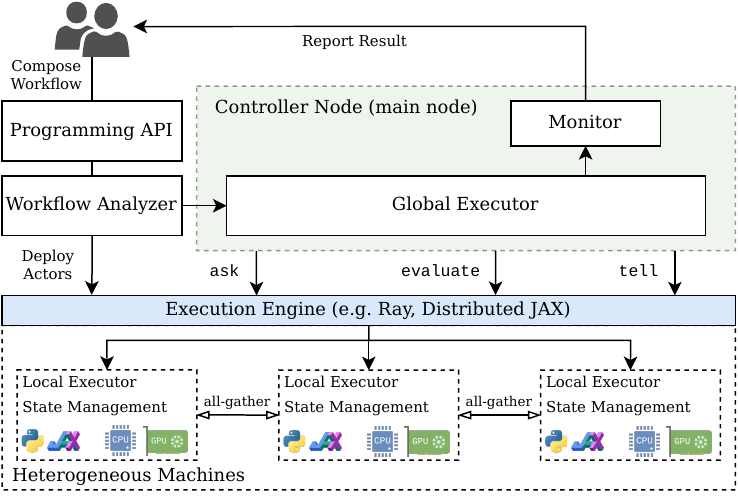}
    \caption{
    Architecture of \ourlib{}. 
    The workflow analyzer sets the stage for task execution. 
    Each node is equipped with a local workflow executor, responsible for orchestrating the \texttt{ask-evaluate-tell} loop.
    At the controller node, a global workflow executor directs the local workflow executors within this loop, employing the \texttt{all-gather} collective operation to harmonize fitness values obtained from the \texttt{evaluate} phase.
    }
    \label{fig:architecture}
\end{figure}

As illustrated by Fig.~\ref{fig:architecture}, the general architecture of \ourlib{} comprises five main components: programming API, workflow analyzer, global workflow running, monitor, and execution engine.
Among them, the programming API and monitor have already been introduced in the previous section. 
In this section, we will first elaborate on the hierarchical state management mechanism, which is the core of the entire architecture.
Then we will delve into the other three main components: workflow analyzer, workflow executor, and execution engine.

\subsection{Hierarchical State Management}

\begin{figure}
    \begin{subfigure}{\columnwidth}
        \includegraphics[width=\columnwidth]{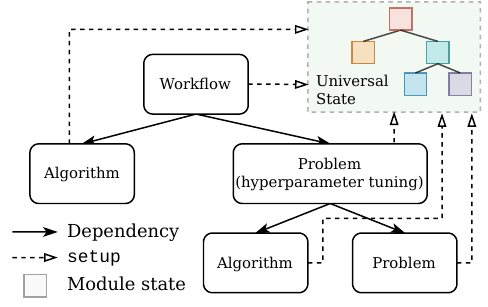}
        \caption{The \emph{universal state} is initialized by traversing the dependency tree recursively and merging states from the current root with those of its children.}
        \label{fig:state_managing_init}
    \end{subfigure}
    
    \begin{subfigure}{\columnwidth}
        \includegraphics[width=\columnwidth]{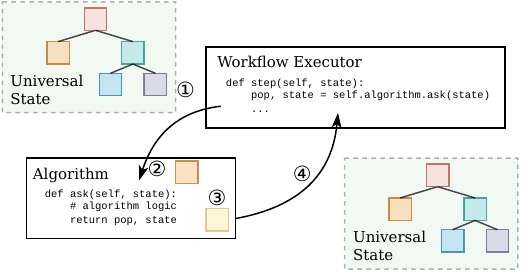}
        \caption{Illustration of state usage at the module level.
        \ding{172} The user provides the universal state.
        \ding{173} The relevant state for the corresponding module is extracted and utilized in the function call.
        \ding{174} The function returns an updated state.
        \ding{175} This updated state is then merged back into the universal state.}
        \label{fig:state_managing_update}
    \end{subfigure}
    \caption{Hierarchical state management in \ourlib{}: (a) state initialization, (b) state update.}
    \label{fig:state_managing}
\end{figure}

As highlighted in the preceding section, \ourlib{} utilizes a \emph{hierarchical state management} mechanism to streamline state transitions throughout the execution process. 
To elucidate this mechanism, let us consider the hyperparameter tuning task illustrated in Fig.~\ref{fig:state_managing}. 
The architecture is structured across three module tiers. 
At the pinnacle is a workflow encapsulating the entire hyperparameter optimization endeavor. 
This workflow branches into two distinct sub-modules: an algorithm module (for optimizing the parameter set) and a problem module (for assessing the performance of a specific parameter set). 
Initially, each module undergoes a recursive state initialization and is assigned a unique identifier. 
When a function accesses the state, it retrieves its own unique identifier and fetches the corresponding state, as depicted in Fig.~\ref{fig:state_managing_init}. 
Post-execution, when the function returns an updated state, this state is seamlessly integrated back into the overarching universal state using the identifier, a process visualized in Fig.~\ref{fig:state_managing_update}.

Although state management introduces additional overhead, it is effectively eliminated by leveraging the JAX's JIT compiler.
The system efficiently resolves the overhead associated with state identification after compilation.
Specifically, JAX first traces the function, converting it into a computational graph where operations are directly linked to the tensors it applies to.
Once this computational graph is compiled, subsequent function calls take advantage of this pre-compiled version, bypassing the need for repeated state identification and ensuring optimal performance. 

Within the proposed hierarchical state management, traditional distinctions between algorithms and problems become nuanced: all elements present themselves as interconnected modules within a comprehensive hierarchical architecture.
This design aligns with the requirement of \emph{Flexibility and Extensibility} as discussed in Section~\ref{sec:motivation}.

\color{black}

\subsection{Workflow Analyzer}

\begin{figure}
    \centering
    \includegraphics[width=\linewidth]{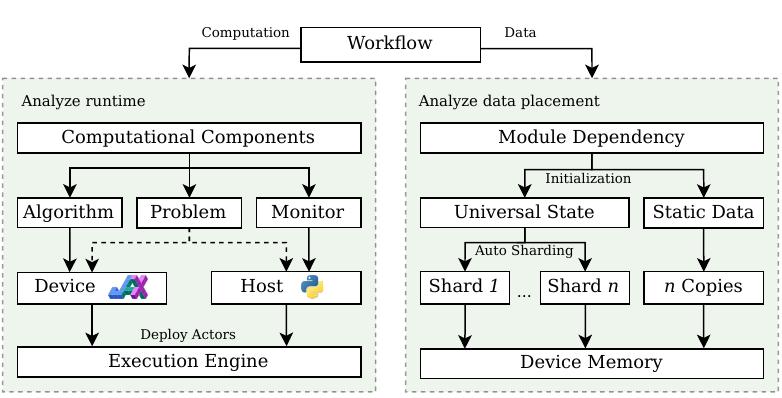}
    \caption{
        Workflow analyzer in \ourlib{}. 
        The analyzer examines both computation and data components distinctly. 
        For computation, it ascertains the optimal runtime: typically, algorithms execute on devices, monitors operate on the host, and the problem can run on either. 
        For data, it strategically distributes data across multiple devices. 
        The universal state is sharded to ensure distribution across all devices, while the static data is replicated consistently to each device.
    }

    \label{fig:workflow-analyze}
\end{figure}

As depicted in Fig.~\ref{fig:workflow-analyze}, the workflow analyzer is instrumental in analyzing and tailoring the execution behavior to align with user requirements. 
Given \ourlib{}'s commitment to the functional programming paradigm, workflows can be modularly decomposed.

In multi-device scenarios, despite the presence of a single host, a multitude of diverse devices can be concurrently harnessed. 
When leveraging multiple local devices, strategic data distribution becomes paramount. 
Primarily, two strategies emerge: \emph{replication} -- duplicating the same data across all devices, and \emph{sharding} -- each device holding only a portion of the complete dataset. 
By aggregating the capabilities of multiple devices, sharding optimizes individual device memory and can handle larger datasets than a standalone device. 
However, it may necessitate data transfer between devices for certain operations, thus introducing potential communication overheads. 
This underscores the need for sharding strategies that mitigate communication costs.

From a data flow perspective, the crux of EC lies in the repetitive update of the fitness and population tensors, $\textbf{T}_{fit}$ and $\textbf{T}_{pop}$, as delineated in Section~\ref{sec:model_computation}. 
Given this, it is imperative that any sharding strategy should consider both tensors.
For $\textbf{T}_{fit}$, it is straightforward that the evaluation of each candidate solution can be performed independently and in parallel.
However, for $\textbf{T}_{pop}$, which represents the tensor of encoded candidate solutions, the situation is more sophisticated.
Basically, there exist two potential sharding strategies: segmenting by individual encoded candidates or by encoded dimensions. 
While the former strategy seems intuitive, it is less effective in the context of EC algorithms, which may frequently require collective information from the entire population for estimation of distribution.
In contrast, sharding by encoded dimension aligns better with the inherent behavior of EC algorithms. During reproductive operations such as crossover or mutation, EC algorithms typically adjust each dimension of the individual candidate solutions in an isolated manner. 
Hence, \ourlib{} adopts \emph{dimension-centric sharding} for $\textbf{T}_{pop}$ to curtail cross-device communication overheads.
It is important to note that while the sharding strategy can be applied universally across various algorithms, its effectiveness can be influenced by specific algorithm designs. 
Particularly, algorithms that engage extensively in cross-dimensional operations may not fully benefit the multi-device acceleration.

In distributed settings, the workflow analyzer synchronizes both the module and its preliminary data across nodes. 
When a sophisticated distributed engine like Ray is employed, the analyzer wraps both the algorithm and problem, including their states, into actor constructs. 
With the execution engine's aid, these actors are dispatched to each node.  
However, when using a basic engine like distributed JAX, the workflow analyzer employs the SPMD (Single Program, Multiple Data) pattern. 
Here, users are tasked with initiating an identical program on all nodes, but with slight input variations (e.g., node IDs). 
This approach ensures natural synchronization during the initialization phase, resulting in uniform module and data distribution across nodes.

Moreover, our workflow analyzer is capable of effectively separating JAX code from standard Python code.
Upon isolation, the JAX code undergoes compilation by XLA, thus facilitating its application across a diverse array of hardware backends such as CPUs, NVIDIA GPUs, and AMD GPUs, among others. 
This adaptability offers users the flexibility to effortlessly transition between execution backends to meet their unique requirements.

\subsection{Workflow Executor}
In multi-device configurations, even after the state partitioning facilitated by the workflow analyzer, the sharding of all intermediate tensors remains paramount. 
To this end, we adopt the GSPMD method~\cite{xu2021gspmd} as embraced by JAX, which ensures consistent sharding propagation for every intermediate tensor. 
Within the realm of distributed computing, our approach focuses on maximizing multi-device capabilities within each node. 
In contrast to specific distributed EC algorithms (e.g., the distributed evolution strategy~\cite{OpenES}), our approach embeds itself at the workflow level. 
This architectural choice endows \ourlib{} with unparalleled adaptability, which enables support for a myriad of algorithms and problems without mandating foundational code alterations.

In detail, the workflow executor disperses the \textbf{Algorithm} and \textbf{Problem} modules as actors across each node while concurrently mirroring the state. 
On one hand, every node employs a local workflow executor tailored for the \texttt{ask-evaluate-tell} loop. 
On the other hand, a centralized global workflow executor oversees the entire node network's execution for coordinating operations across local executors. 
Central to its approach is the \texttt{all-gather} collective operation which synchronizes fitness values returned by the \texttt{evaluate} function. 
This synchronized aggregation paves the way for a unified update during the \texttt{tell} step, thus ensuring end-to-end synchronization at the end of each iteration.

\subsection{Execution Engine}
The execution engine refers to the component specially engineered to coordinate and synchronize the execution of code across disparate machines.
Within the architecture of \ourlib{}, we predominantly leverage two salient execution engines: Ray and distributed JAX.
As a Python-centric framework, Ray is underpinned by an actor-based programming paradigm. 
Users can delegate actors to Ray's scheduler, which then automatically allocates these actors to machines in accordance with stipulated computational requirements. 
In contrast, the distributed functionality within JAX deviates from that of Ray. 
Rather than incorporating scheduling capabilities, this feature of JAX prioritizes synchronization tools and facilitates collective operations across a multitude of nodes.

\section{Implementation} \label{sec:implementation}

\begin{table}[]
    \centering
    \footnotesize
    \caption{Selected EC algorithms in \ourlib{} for single-objective optimization: Evolution Strategy (ES), Particle Swarm Optimization (PSO), and Differential Evolution (DE).}
    \begin{tabular}{cc}
    \toprule
    Type & Algorithm Name \\ 
    \midrule
    ES & \begin{tabular}[c]{@{}c@{}} CMA-ES~\cite{CMAES}, PGPE~\cite{PGPE}, OpenES~\cite{OpenES}, \\  CR-FM-NES~\cite{CR_FM_NES}, xNES~\cite{xNES}, ...\end{tabular} \\ 
    \midrule
    PSO & \begin{tabular}[c]{@{}c@{}}FIPS~\cite{FIPS}, CSO~\cite{CSO}, CPSO~\cite{CPSO},\\ CLPSO~\cite{CLPSO}, SL-PSO~\cite{SLPSO}, ...\end{tabular} \\ 
    \midrule
    DE & \begin{tabular}[c]{@{}c@{}}CoDE~\cite{CoDE}, JaDE~\cite{JADE}, SaDE~\cite{SaDE},\\ SHADE~\cite{SHADE}, IMODE~\cite{IMODE}, ...\end{tabular} \\ 
    \bottomrule
\end{tabular}
    \label{tab:algorithm_list_so}
\end{table}

\begin{table}[]
    \centering
    \caption{Selected EC algorithms in \ourlib{} for multi-objective optimization: dominance-based, decomposition-based, and indicator-based approaches.}
    \begin{tabular}{cc}
    \toprule
    Type & Algorithm Name \\ 
    \midrule
    Dominance-based & \begin{tabular}[c]{@{}c@{}}NSGA-II~\cite{NSGA-II}, NSGA-III~\cite{NSGA-III}, SPEA2~\cite{SPEA2},
    \\ BiGE\cite{BiGE}, KnEA~\cite{KnEA}, ...\end{tabular} \\ 
    \midrule
    Decomposition-based & \begin{tabular}[c]{@{}c@{}}MOEA/D~\cite{MOEAD}, RVEA~\cite{RVEA}, t-DEA~\cite{tDEA},
    \\ MOEAD-M2M~\cite{MOEADM2M}, EAG-MOEAD~\cite{EAGMOEAD}, ...\end{tabular} \\ 
    \midrule
    Indicator-based & \begin{tabular}[c]{@{}c@{}}IBEA~\cite{IBEA}, HypE~\cite{HypE}, SRA~\cite{SRA},
    \\  MaOEA-IGD~\cite{MaOEA-IGD}, AR-MOEA~\cite{ARMOEA}, ...\end{tabular} \\ 
    \bottomrule
\end{tabular}
    \label{tab:algorithm_list_mo}
\end{table}

\begin{table}[]
\centering
\caption{Benchmark problems provided by \ourlib{}.}
\begin{tabular}{M{2cm} M{6cm}} 
\toprule
Name & Description \\
\midrule
CEC'22~\cite{cec22} & A test suite for benchmarking single-objective numerical optimization. \\ 
\midrule
ZDT~\cite{ZDT} & A test suite for benchmarking bi-objective numerical optimization. \\
\midrule
DTLZ~\cite{DTLZ} & A test suite for benchmarking multi-objective numerical optimization with scalable dimensions. \\
\midrule
MaF~\cite{MaF} & A test suite for benchmarking multi-objective numerical optimization with many objectives, a.k.a., many-objective optimization. \\
\midrule
LSMOP~\cite{LSMOP} & A test suite for benchmarking multi-objective numerical optimization with large-scale decision variables. \\
\midrule
Brax~\cite{brax} & A JAX-based physics engine, fully differentiable and optimized for reinforcement learning, robotics, and other simulation-intensive applications. \\
\midrule
Gym~\cite{openai_gym} & A comprehensive toolkit for developing and comparing reinforcement learning algorithms through a collection of standardized environments. \\
\bottomrule
\end{tabular}
\label{tab:problem_list}
\end{table}

\ourlib{} is developed in Python and leverages JAX for optimized execution on hardware accelerators. 
The design integrates distributed computing features using either Ray or distributed JAX. 
For the implementation using Ray, modules and their respective states are treated as actors, and assigned to Ray's scheduler. 
For the implementation using distributed JAX, users are asked to initiate a consistent program across nodes, but with distinctions in input arguments. 
Based on this foundation, we have further crafted a comprehensive library.
For user convenience, \ourlib{} is readily available on PyPI, facilitating an effortless installation process via the \texttt{pip} command. 
Moreover, we provide a comprehensive documentation\footnote{The documentation of \ourlib{} is available at: \url{https://evox.readthedocs.io/}.} for detailed user guidance.

In detail, for single-objective optimization, \ourlib{} presents EC algorithms spanning categories such as Evolution Strategy (ES), Particle Swarm Optimization (PSO), and Differential Evolution (DE), as listed in Table~\ref{tab:algorithm_list_so}. 
for multi-objective optimization, the library covers all three representative types of EC algorithms: dominance-based, decomposition-based, and indicator-based ones, as listed in Table~\ref{tab:algorithm_list_mo}. 
Beyond algorithms, \ourlib{} also provides seamless support for a diverse set of benchmark problems, which fulfills various research and application requirements. 
This collection includes benchmark test suites for numerical optimization and environments for reinforcement learning tasks, as listed in Table~\ref{tab:problem_list}.
All algorithms and problems are implemented in a tensorized manner to maximize the parallelization benefits of GPU acceleration.

\section{Experimental Study\label{sec:experiments}}
This section offers a comprehensive assessment of the performance of \ourlib{} through a series of experiments. 
Except for the multi-node acceleration experiment presented in Section~\ref{sec:exp_2_3}, all the other experiments were executed on a dedicated machine equipped with two Intel Xeon Gold 6226R CPU @ 2.90GHz processors (each with 16 cores and 32 threads)\footnote{For experiments running on CPUs, parallelization was optimized across all $2\times32$ threads.} and eight NVIDIA A100 GPUs, with only one GPU engaged for each experiment.
In contrast, the experiment in Section~\ref{sec:exp_2_3} utilized a 4-node physical GPU cluster interconnected using 10Gb Ethernet over optical fiber, where each node integrated an Intel Xeon Gold 6240 CPU @ 2.60GHz and four NVIDIA RTX 2080ti GPUs. 
The pseudo-random number generation remained consistent, employing the \texttt{rbg} generator from JAX.

\subsection{System Performance}
\label{sec:exp_1}

The primary objective of this subsection is to evaluate the efficacy of GPU acceleration in handling both single-objective and multi-objective numerical optimization tasks. 
Following this, we scrutinize the framework's proficiency in multi-node acceleration.

\subsubsection{Single-objective Numerical Optimization}
\label{sec:exp_2_1}

In this experiment, we evaluate three representative EC algorithms (PSO~\cite{kennedy_particle_1995}, DE~\cite{storn_differential_1997}, and CMA-ES~\cite{CMAES}) on the Sphere function, which is a basic benchmark function for single-objective numerical optimization.
During the assessment of scaling with respect to the problem dimension, we retained a consistent population size of 100. 
On the other hand, in the evaluation of population size scalability, we maintained the problem dimension at a constant value of 100.

\begin{figure}
    \centering
    \begin{subfigure}{0.45\columnwidth}
        \centering
        \includegraphics[width=\columnwidth]{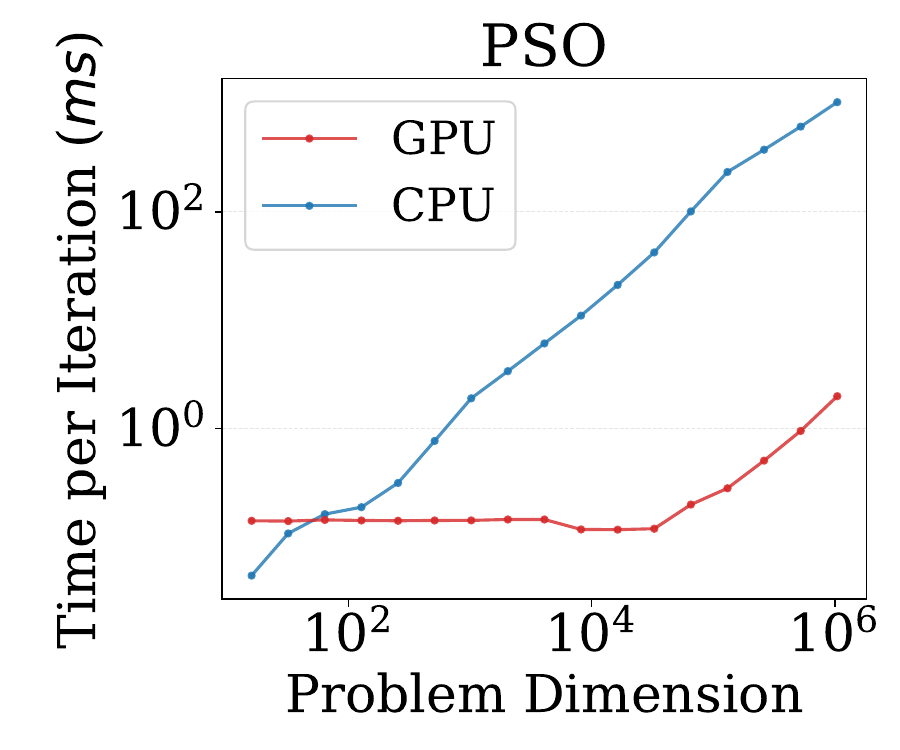}
    \end{subfigure}
    \begin{subfigure}{0.45\columnwidth}
        \centering
        \includegraphics[width=\columnwidth]{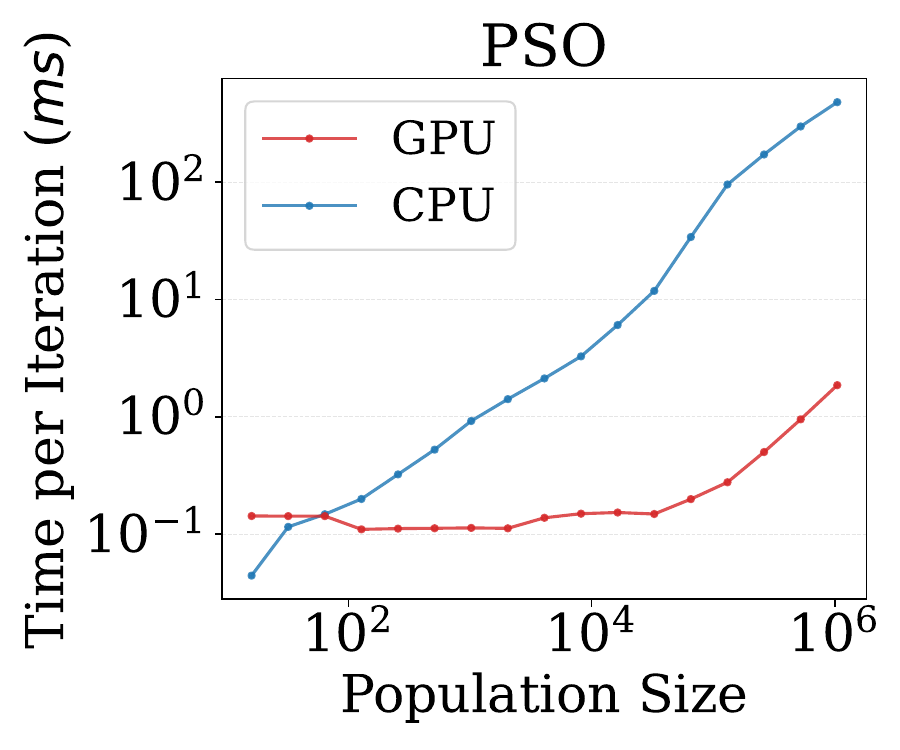}
    \end{subfigure}

    \begin{subfigure}{0.45\columnwidth}
        \centering
        \includegraphics[width=\columnwidth]{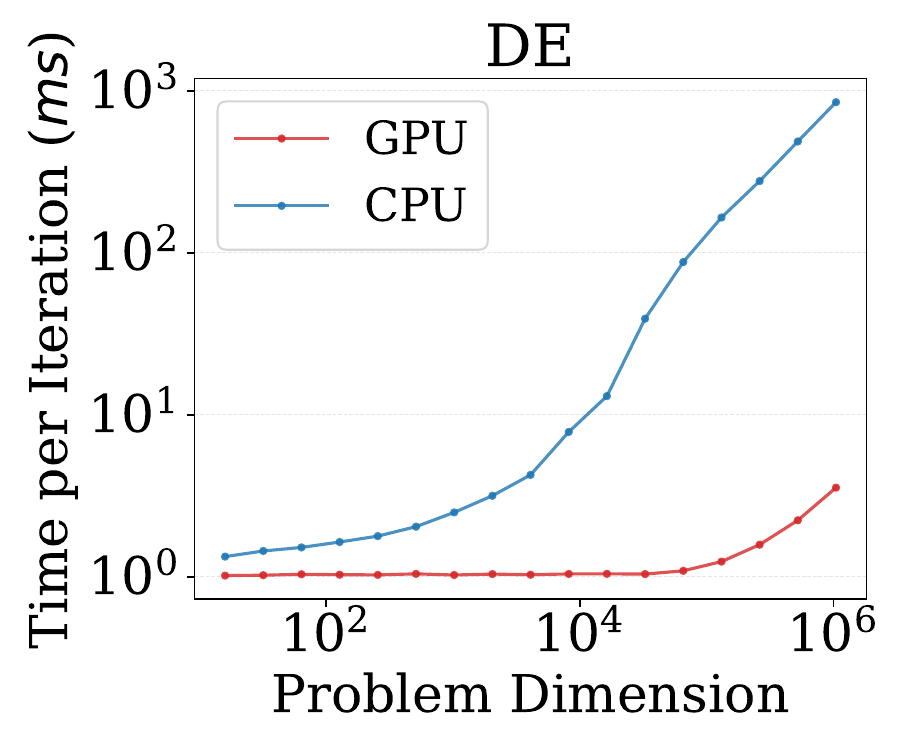}
    \end{subfigure}
    \begin{subfigure}{0.45\columnwidth}
        \centering
        \includegraphics[width=\columnwidth]{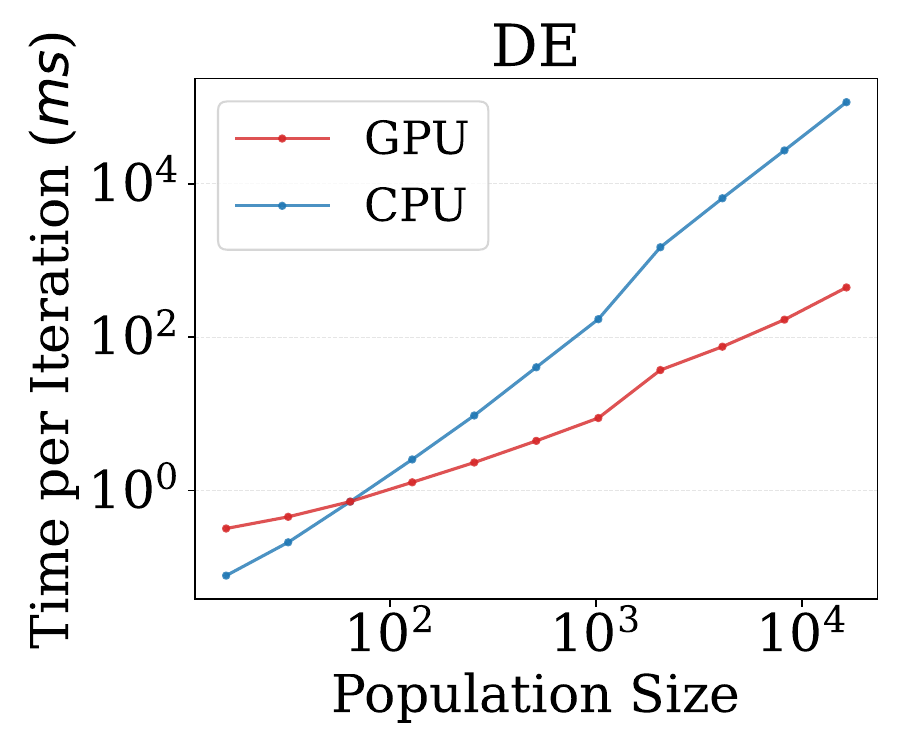}
    \end{subfigure}

    \begin{subfigure}{0.45\columnwidth}
        \centering
        \includegraphics[width=\columnwidth]{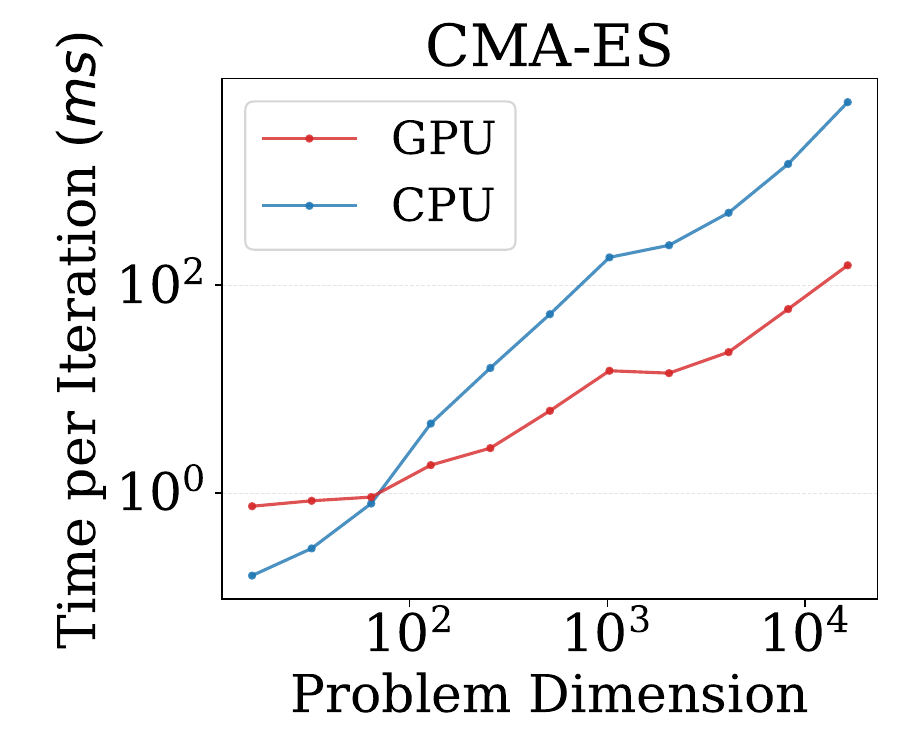}
    \end{subfigure}
    \begin{subfigure}{0.45\columnwidth}
        \centering
        \includegraphics[width=\columnwidth]{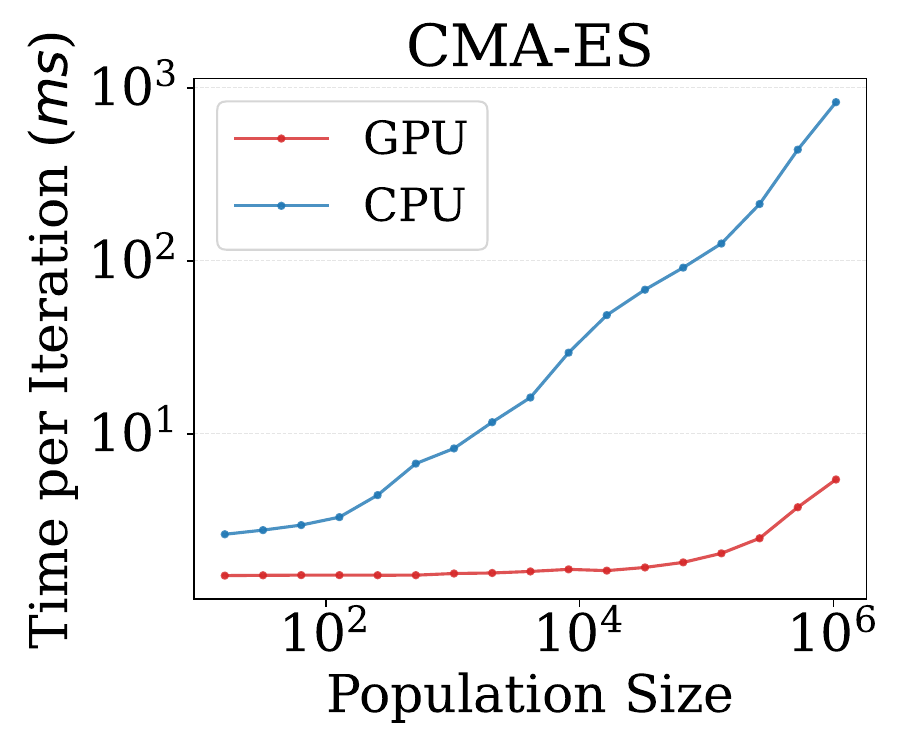}
    \end{subfigure}
    \caption{
    Scalability performance of PSO, DE, and CMA-ES on the Sphere function for single-objective numerical optimization, in terms of problem dimension and population size. Both axes employ a logarithmic scale.
    A fixed population size of 100 is used when scaling the problem dimension, and vice versa.}
    \label{fig:system_performance_single}
\end{figure}

Fig.~\ref{fig:system_performance_single} reveals that GPU acceleration considerably enhances the performance of the evaluated algorithms, especially as the problem's dimension or population size grows.
Initial tests with small dimensions or populations might favor the CPU, but GPU performance rapidly overtakes as the scale increases, frequently achieving a tenfold or greater speedup.
A notable observation is the plateauing of performance in the early stages of the scaling tests with GPU acceleration. 
This plateau suggests that lighter computations cannot fully harness the GPU's capabilities, thus resulting in near-constant computational costs.

However, it is paramount to understand that the advantages of GPU acceleration are algorithm-dependent. 
Algorithms inherently unsuited for parallelism or those restricted by memory constraints might not benefit as significantly. 
For instance, CMA-ES internally uses a covariance matrix, demanding memory proportional to the square of the problem dimension. 
This requirement limited its ability to scale beyond a dimension of 16,384, even though the GPU accelerated performance by orders of magnitude. 
The vanilla DE also presents some limitations, particularly when scaling the population size beyond 16,384. 
Such restrictions emerge from specific operators within DE not optimized for GPUs or larger populations. 
Specifically, the mutation operator in vanilla DE, while suitable for smaller populations, becomes computationally intensive with larger populations, especially when ensuring distinct individual sampling.

\subsubsection{Multi-objective Numerical Optimization}
\label{sec:exp_2_2}

This subsection evaluates the advantages of GPU acceleration for multi-objective EC algorithms, segmented into two primary investigative parts.
First, we investigated the scalability of three representative multi-objective EC algorithms: NSGA-II~\cite{NSGA-II}, MOEA/D~\cite{MOEAD}, and IBEA~\cite{IBEA}. 
When scaling the problem dimension, we consistently employed a population size of 100, and vice versa.
Subsequently, we assessed the scalability of another three representative multi-objective EC algorithms: NSGA-III~\cite{NSGA-III}, RVEA\cite{RVEA}, and HypE~\cite{HypE} in relation to the number of optimization objectives. 
For this part, the problem dimension was fixed at 100,000. 
All experiments were conducted using DTLZ1~\cite{DTLZ}, one of the most commonly used test functions for multi-objective numerical optimization.

\begin{figure}
    \centering
    \begin{subfigure}{0.45\columnwidth}
        \centering
        \includegraphics[width=\columnwidth]{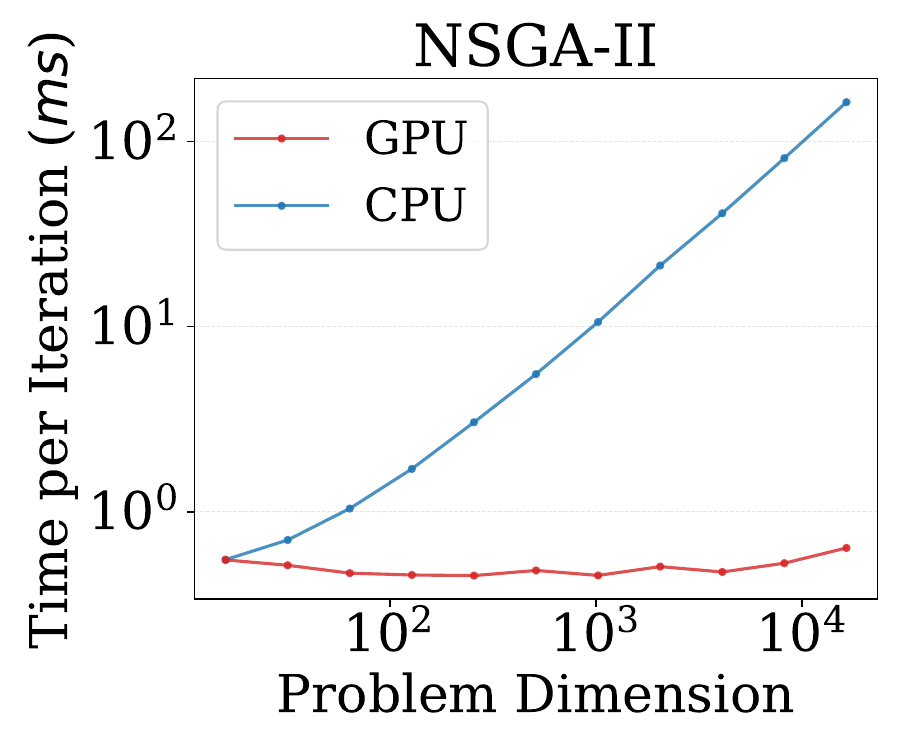}
    \end{subfigure}
    \begin{subfigure}{0.45\columnwidth}
        \centering
        \includegraphics[width=\columnwidth]{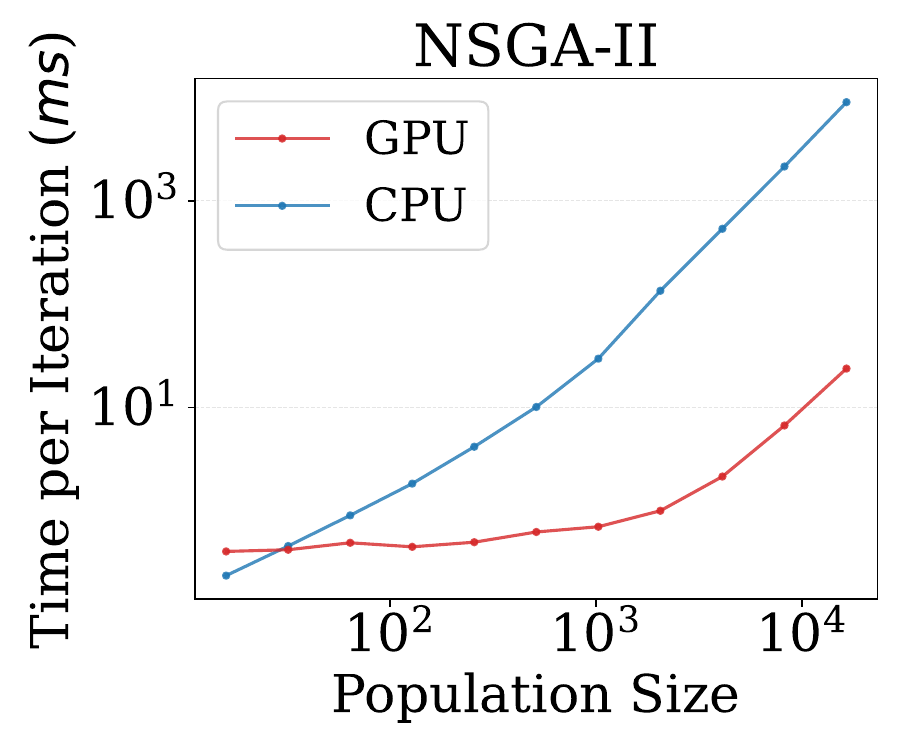}
    \end{subfigure}
    \begin{subfigure}{0.45\columnwidth}
        \centering
        \includegraphics[width=\columnwidth]{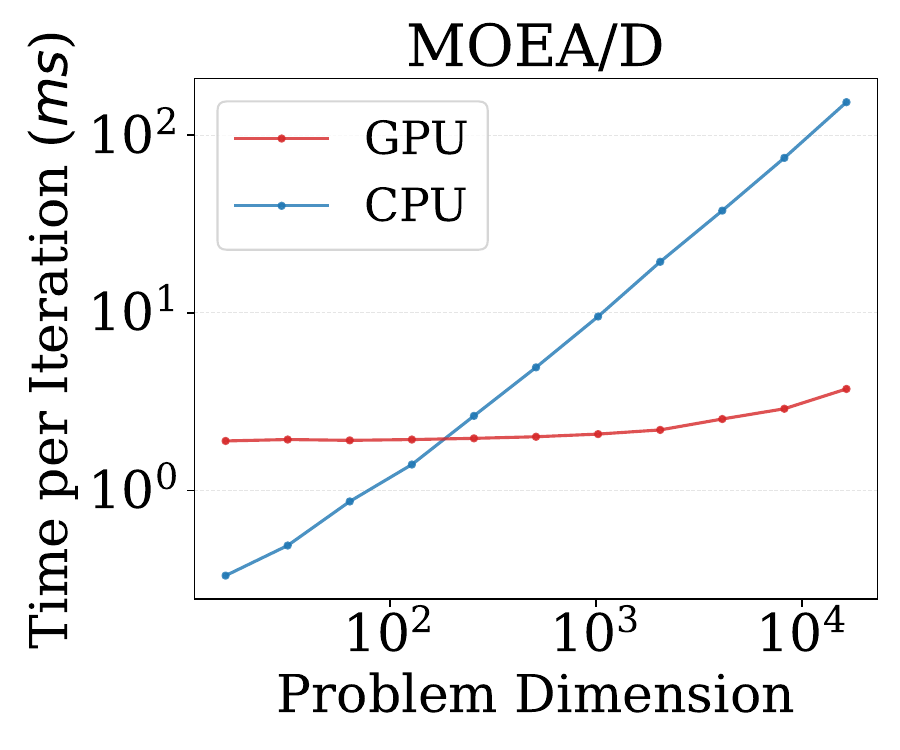}
    \end{subfigure}
    \begin{subfigure}{0.45\columnwidth}
        \centering
        \includegraphics[width=\columnwidth]{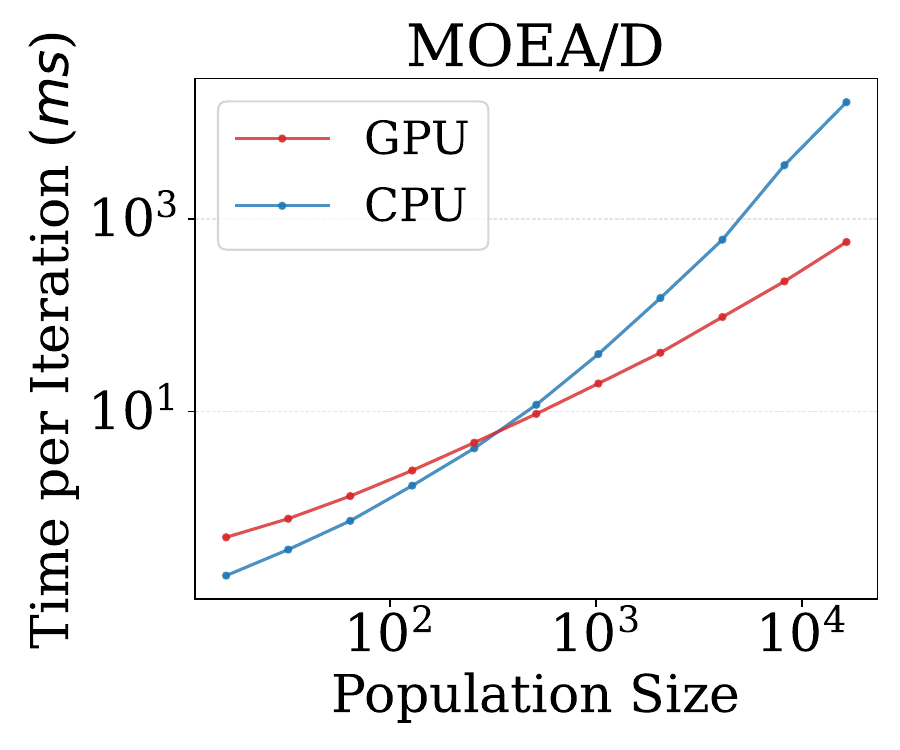}
    \end{subfigure}
    \begin{subfigure}{0.45\columnwidth}
        \centering
        \includegraphics[width=\columnwidth]{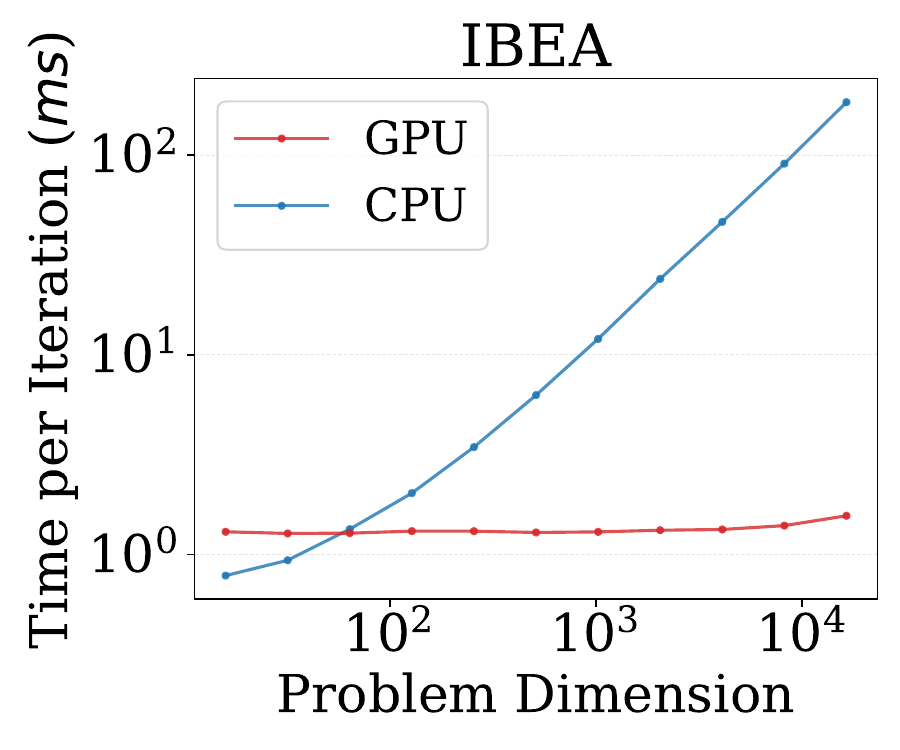}
    \end{subfigure}
    \begin{subfigure}{0.45\columnwidth}
        \centering
        \includegraphics[width=\columnwidth]{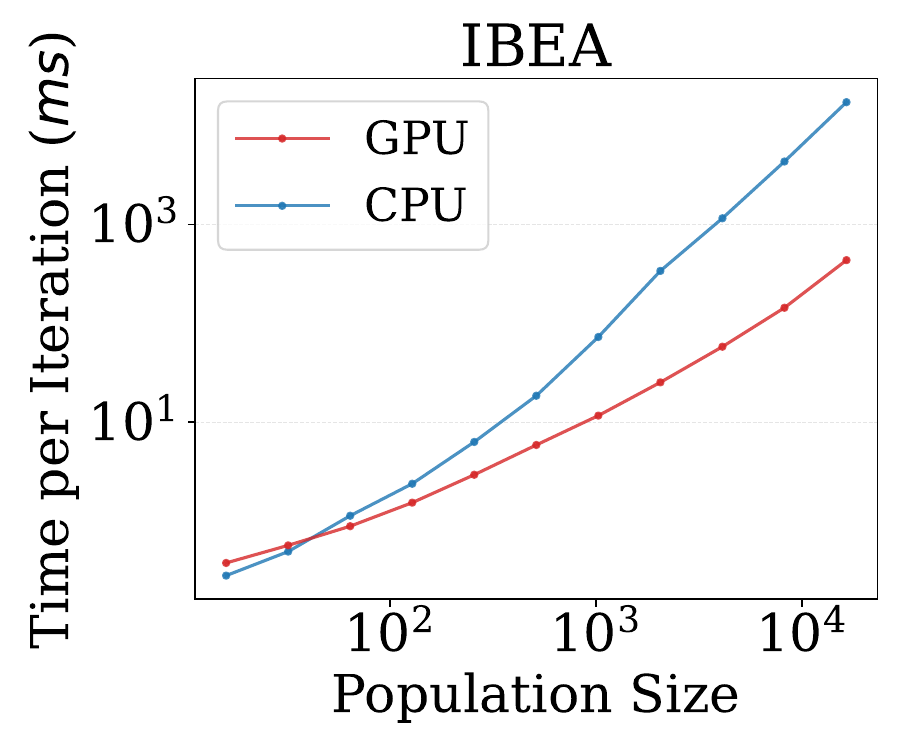}
    \end{subfigure}
    \caption{Scalability performance of NSGA-II, MOEA/D, and IBEA on the DTLZ1 function for multi-objective numerical optimization, in terms of problem dimension and population size. 
    Both axes employ a logarithmic scale.
    A fixed population size of 100 is used when scaling the problem dimension, and vice versa.
    }
    \label{fig:system_performance_multi}
\end{figure}

As shown in Fig.~\ref{fig:system_performance_multi}, NSGA-II notably benefits from GPU acceleration during both problem dimension and population size scaling. 
While MOEA/D's scalability is not as pronounced as NSGA-II, it still substantially benefits from GPU acceleration. 
IBEA also demonstrates significant speed enhancements, particularly when the problem dimension is high.

Notably, the effectiveness of GPU acceleration is intrinsically related to the algorithmic mechanisms. 
For example, the performance improvement in NSGA-II is attributed to the GPU-accelerated computation of dominance relations, which is computationally intensive but amenable to parallelization. 
In contrast, the scalability of MOEA/D is somehow limited by its inherent design.
Specifically, the sequential update strategy within the reproduction operator requires the completion of one individual's update before proceeding to the next.
This sequential dependency hampers the potential for parallel processing, which is crucial for GPU acceleration.

\begin{figure}
    \centering
    \begin{subfigure}{0.31\columnwidth}
        \centering
        \includegraphics[width=\columnwidth]{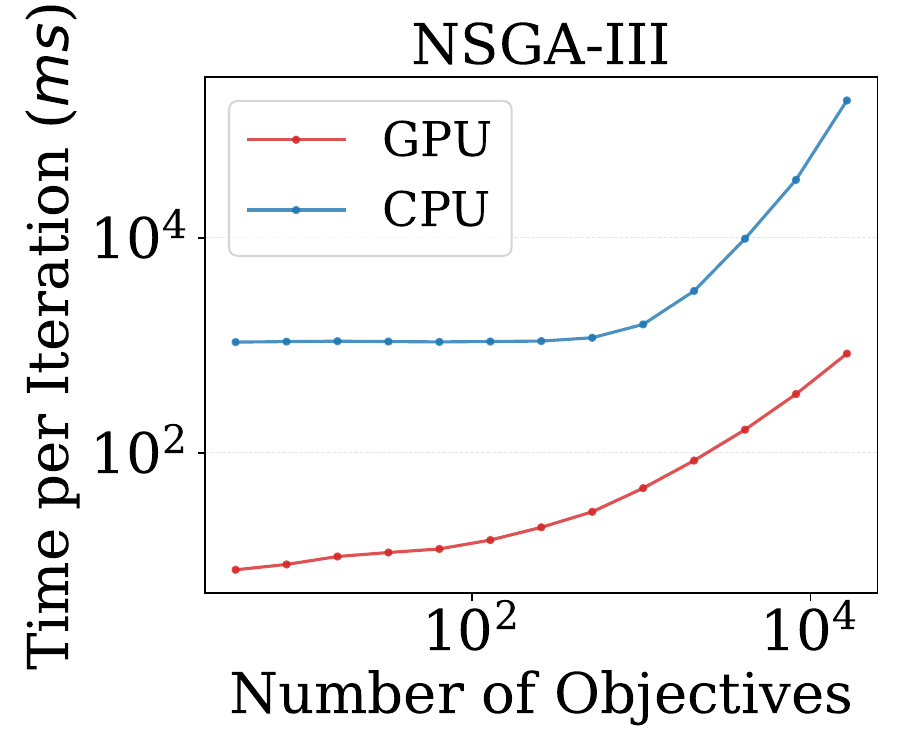}
    \end{subfigure}
    \begin{subfigure}{0.31\columnwidth}
        \centering
        \includegraphics[width=\columnwidth]{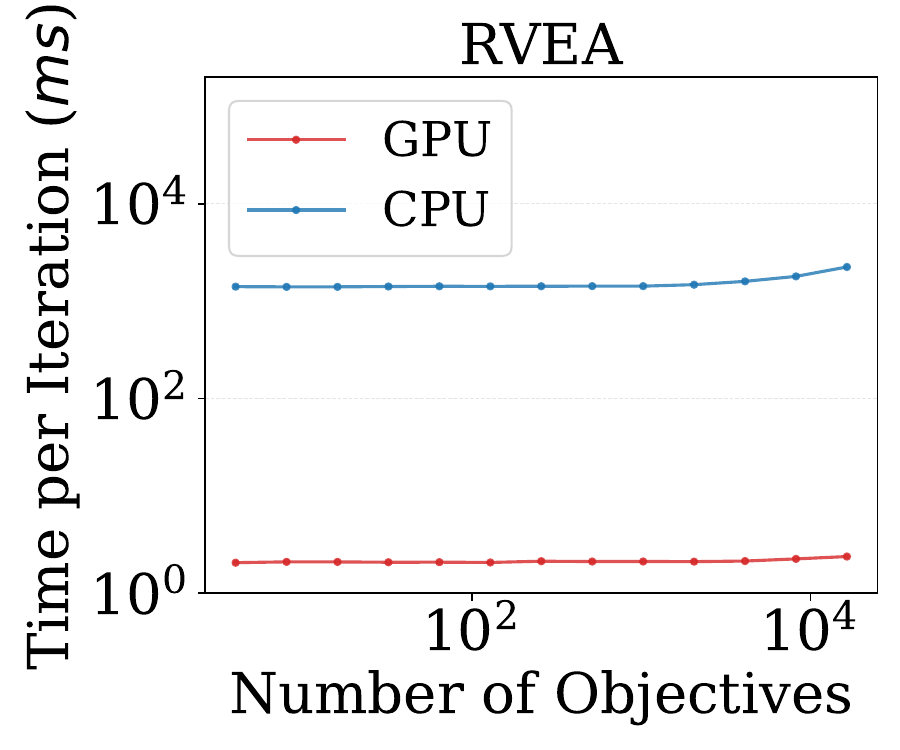}
    \end{subfigure}
    \begin{subfigure}{0.31\columnwidth}
        \centering
        \includegraphics[width=\columnwidth]{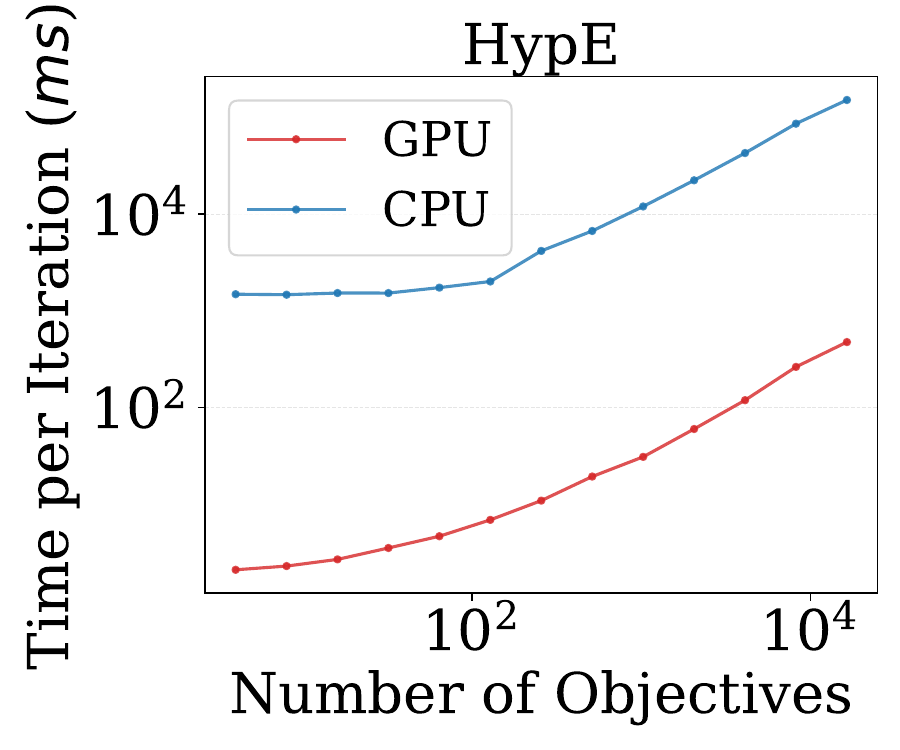}
    \end{subfigure}
    \caption{Scalability performance of NSGA-III, RVEA, and HypE on the DTLZ1 function for multi-objective numerical optimization, in terms of the number of optimization objectives.
    Both axes employ a logarithmic scale.
    The problem dimension is fixed at 100,000.
    }
    \label{fig:system_performance_scale_obj}
\end{figure}

Fig.~\ref{fig:system_performance_scale_obj} indicates that all tested algorithms significantly benefit from GPU acceleration as the number of optimization objectives increases. 
Although these algorithms were not inherently designed for a large number of optimization objectives, they maintain consistent performance up to 100 objectives. 
However, as the count escalates, NSGA-III and HypE's performance diminishes while RVEA remains consistently robust, although the scenarios involving over 100 optimization objectives are rare in practice.

\begin{table}
    \centering
    \footnotesize
    \caption{Architecture of the CNN used in the multi-node acceleration experiment.}
    \begin{tabular}{cccc}
        \toprule
        Input Shape & Layer & Filter Shape & Strides \\
        \toprule
        $32\times 32 \times 3$ & Conv & $3\times 3\times 3 \times 32$ & 1\\
        $30\times 30 \times 32$ & Max Pooling & $2\times 2$ & 2\\
        $15\times 15 \times 32$ & Conv & $3\times 3\times 32 \times 32$ & 1\\
        $13 \times 13 \times 32$ & Max Pooling & $2\times 2$ & 2\\
        $6 \times 6 \times 32$ & Conv & $3\times 3\times 32 \times 32$ & 1\\
        $512$ & Fully Connected & $512\times 64$ & ---\\
        $64$ & Fully Connected & $64\times 10$ & ---\\
        \bottomrule
    \end{tabular}

    \label{tab:arch}
\end{table}

\subsubsection{Multi-node Acceleration}
\label{sec:exp_2_3}
This subsection evaluates the efficacy of multi-node acceleration and contrasts the performance between the two execution engines leveraging JAX and Ray respectively. 
For this purpose, we conducted an experiment on neuroevolution for image classification across multiple GPU devices.

Specifically, we evolved a convolutional neural network (CNN) on the CIFAR-10 dataset~\cite{cifar10}, scaling from 4 to 16 GPUs, and measured the time per iteration. 
The CNN architecture, as detailed in Table~\ref{tab:arch}, employs ReLU~\cite{nair_relu_2010} as the activation function between layers. 
For the EC algorithm, we utilized PGPE~\cite{sehnke_parameter-exploring_2010} with a population size of 192.
To quantify the acceleration's effectiveness, we present two metrics: time per iteration and relative performance, the latter being the inverse of the former.

\begin{figure}
    \centering
    \begin{subfigure}{0.464\columnwidth}
        \centering
        \includegraphics[width=\columnwidth]{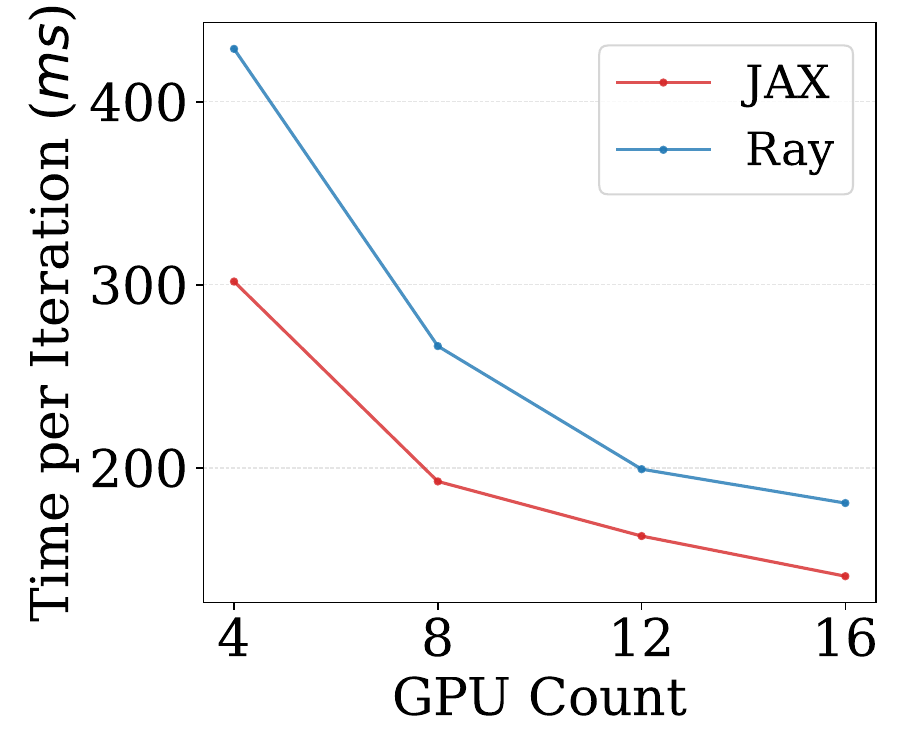}
    \end{subfigure}
    \begin{subfigure}{0.436\columnwidth}
        \centering
        \includegraphics[width=\columnwidth]{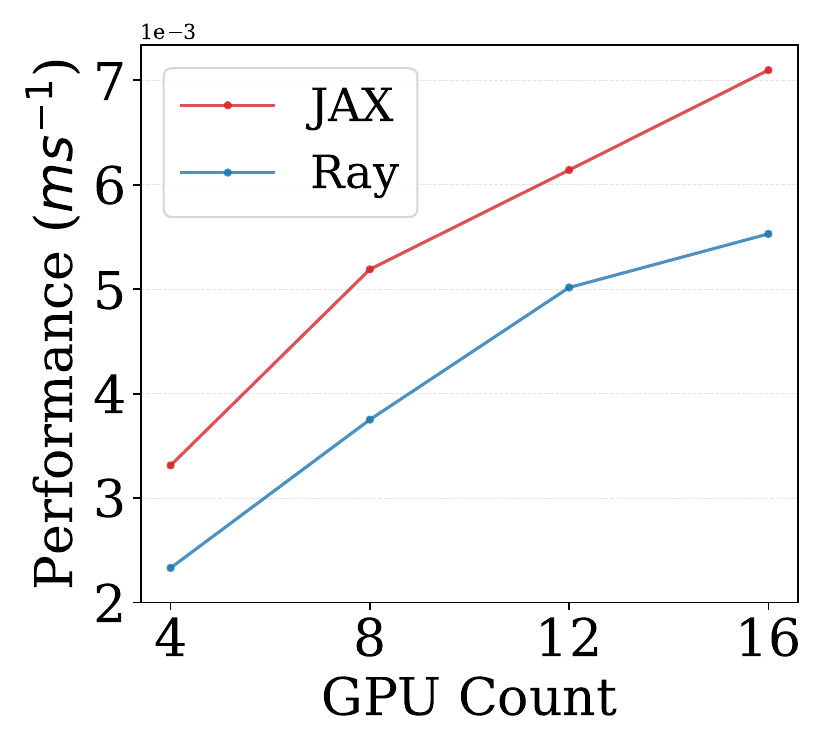}
    \end{subfigure}
    \caption{Results of multi-node acceleration using 4 to 16 GPUs. 
    Data is presented in two forms: runtime (left) and its inverse (right).}
    \label{fig:acceleration_performance}
\end{figure}

Fig.~\ref{fig:acceleration_performance} presents the performance of multi-node acceleration. 
Notably, for a lower count of GPU nodes, the runtime substantially reduces with increased GPU nodes, achieving an almost linear acceleration rate. 
However, as more GPUs are integrated, the rate of performance gain tapers off, leading to an overarching sub-linear acceleration trend.
This behavior aligns with expectations. 
While our acceleration framework primarily accelerates the computational parts of the workflow, the distributed execution engine introduces some overhead. 
As GPU count escalates, the cost of fitness evaluations drops, thus making other workflow costs more prominent.

Notably, the scalability of our distributed workflow is closely tied to the problem's nature.
More computationally demanding problems offer better scalability since the distributed framework effectively offloads the fitness evaluations, thus yielding substantial gains. 
By contrast, for computationally cheap problems (e.g., numerical optimization), the performance enhancements can be less significant.
This is because the algorithm's demands and the distributed framework's overhead often outweigh the computational cost of fitness evaluations.

Besides, the two execution engines leveraging JAX and Ray possess unique performance attributes respectively. 
JAX incurs relatively lesser overhead and offers superior scalability, significantly outperforming Ray on a 4-GPU setup. 
This performance disparity stems from JAX's efficient hardware utilization, especially when recognizing that the 4 GPUs reside on the same physical node. 
By contrast, Ray offers a more intuitive interface, supplemented by features like scheduling and fault tolerance, which are capabilities absent in distributed JAX. 
Ray's scheduling allows users to initiate a task once, distributing it automatically across nodes, while JAX mandates manual task initiation on each node.

\subsection{Model Performance}
\label{sec:exp_4}
Within \ourlib{}, we have seamlessly integrated a range of black-box optimization challenges into the \textbf{Problem} module, all adhering to a unified interface. 
Among these, the reinforcement learning tasks stand out as particularly intricate. 
To assess the model performance of \ourlib{}, we present two distinct demonstrations: one leveraging the CPU-centric Gym~\cite{openai_gym} platform, and the other utilizing the GPU-accelerated Brax~\cite{brax} platform. 
In both cases, the \textbf{Problem} module proficiently manages the interaction between the policy models and the reinforcement learning environments, enabling the EC algorithm to singularly concentrate on refining the policy model's weights through neuroevolution, independent of the specificities of the task at hand.
For a comprehensive evaluation of \ourlib{}'s capabilities, we benchmarked an ES algorithm (PGPE~\cite{PGPE}), in comparison with the widely acknowledged baseline (PPO2 baseline~\cite{schulman2017ppo2}) as endorsed by OpenAI~\cite{OpenAI_rl_baselines}.

\subsubsection{Performance with Gym}
\begin{listing}
\begin{minted}[
    fontsize=\footnotesize,
    linenos,
    frame=lines,
    framesep=2mm,
    xleftmargin=6pt,
    autogobble,
    numbersep=2pt
]{python}
problem = Gym(
    env_name=..., # Gym's environment name
    policy=..., # your policy
    num_workers=..., # number of CPU workers
    env_per_worker=..., # environments per worker
)
\end{minted}
    \caption{
    Configuration for setting up a Gym-based reinforcement learning task in \ourlib{}. 
    Users simply need to specify the environment, define the policy model, and adjust runtime parameters to optimize CPU utilization.
    }
\label{lst:rl_gym_example}
\end{listing}

Over the years, Gym has emerged as an essential open-source platform for developing and benchmarking reinforcement learning algorithms. 
It offers a plethora of predefined environments, streamlining the testing and comparison of various algorithms on a standardized platform. 
In 2021, the development of Gym transitioned to Gymnasium~\cite{towers_gymnasium_2023}, serving as a direct replacement.

As illustrated in Lst.~\ref{lst:rl_gym_example}, setting up a Gym-based problem in \ourlib{} is straightforward. 
Users simply specify a Gym-supported environment and define the policy network's forward function. 
This function primarily accepts two inputs: the network's weight and the observational data from the environment.  
The EC algorithm outputs a varied set of weights for the policy network, which are then evaluated within the specified Gym environment to aggregate rewards (i.e., fitness values).
Additionally, the runtime configuration can be tailored to best align with users' computational resources and needs.

\begin{figure}
    \centering
    \includegraphics[width=\columnwidth]{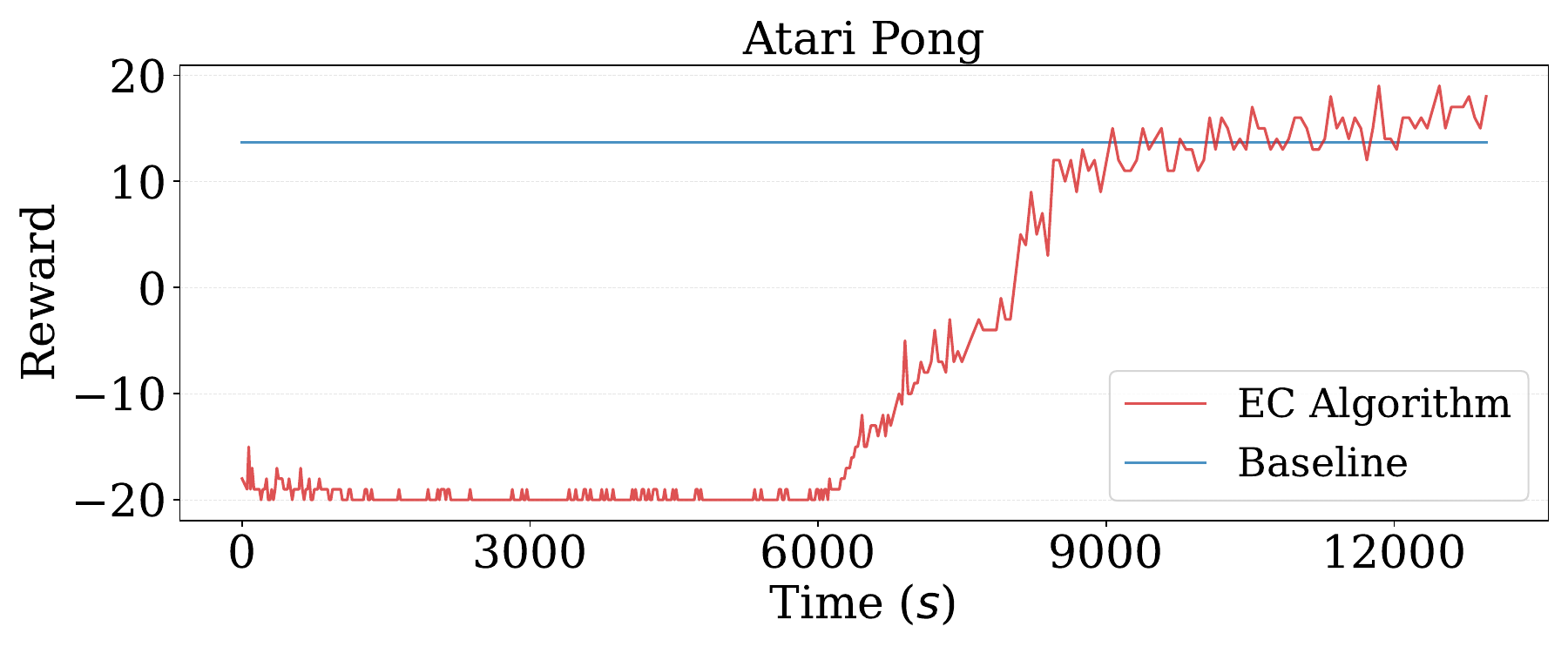}
    \caption{
    Performance curve when tackling the Atari Pong task with Gym. 
    The EC algorithm is PGPE with a population size of 256.
    The baseline performance, achieved by PPO2, was directly taken from \cite{OpenAI_rl_baselines} as a reference.
    }
    \label{fig:rl_gym}
\end{figure}

Specifically, we instantiated an Atari Pong task with the policy model being a CNN with 78,102 parameters.
As shown in Fig.~\ref{fig:rl_gym}, the complexity of Atari Pong and the constraints of the CPU-centric game emulator significantly affected the speed of the workflow. 
Nonetheless, thanks to \ourlib{}'s efficient architecture, all available CPU cores were maximized, accomplishing the tasks in roughly 4 hours to reach the baseline performance.

\subsubsection{Performance with Brax}

\begin{listing}
\begin{minted}[
    fontsize=\footnotesize,
    linenos,
    frame=lines,
    framesep=2mm,
    xleftmargin=6pt,
    autogobble,
    numbersep=2pt
]{python}
problem = Brax(
    env_name=..., # Brax's environment name
    policy=..., # user's policy model
    batch_size=..., # concurrency of environments
)
\end{minted}
\caption{
Configuration for setting up a Brax-based reinforcement learning task in \ourlib{}. 
Users simply need to specify the environment, define the policy model, and determine the \texttt{batch\_size} for specifying the number of environments running in concurrency.
}
\label{lst:rl_brax_example}
\end{listing}

Brax is a differentiable physics engine developed in JAX, which capitalizes on JAX's capabilities to harness GPUs for extensive parallel simulations. 
Given that \ourlib{} shares its foundation with JAX, it seamlessly integrates with Brax.

As illustrated in Lst.~\ref{lst:rl_gym_example}, setting up a Brax-centric problem in \ourlib{} is similar to the case with Gym, necessitating the environment's name and the forward function for the policy network. 
A unique aspect of Brax is the \texttt{batch\_size} parameter, which indicates the number of concurrent environments on the hardware accelerator. 
This often aligns with the population size of an EC algorithm to harness the prowess of Brax by batch-evaluating the environments on GPU(s).

\begin{figure}
    \centering
    \includegraphics[width=\columnwidth]{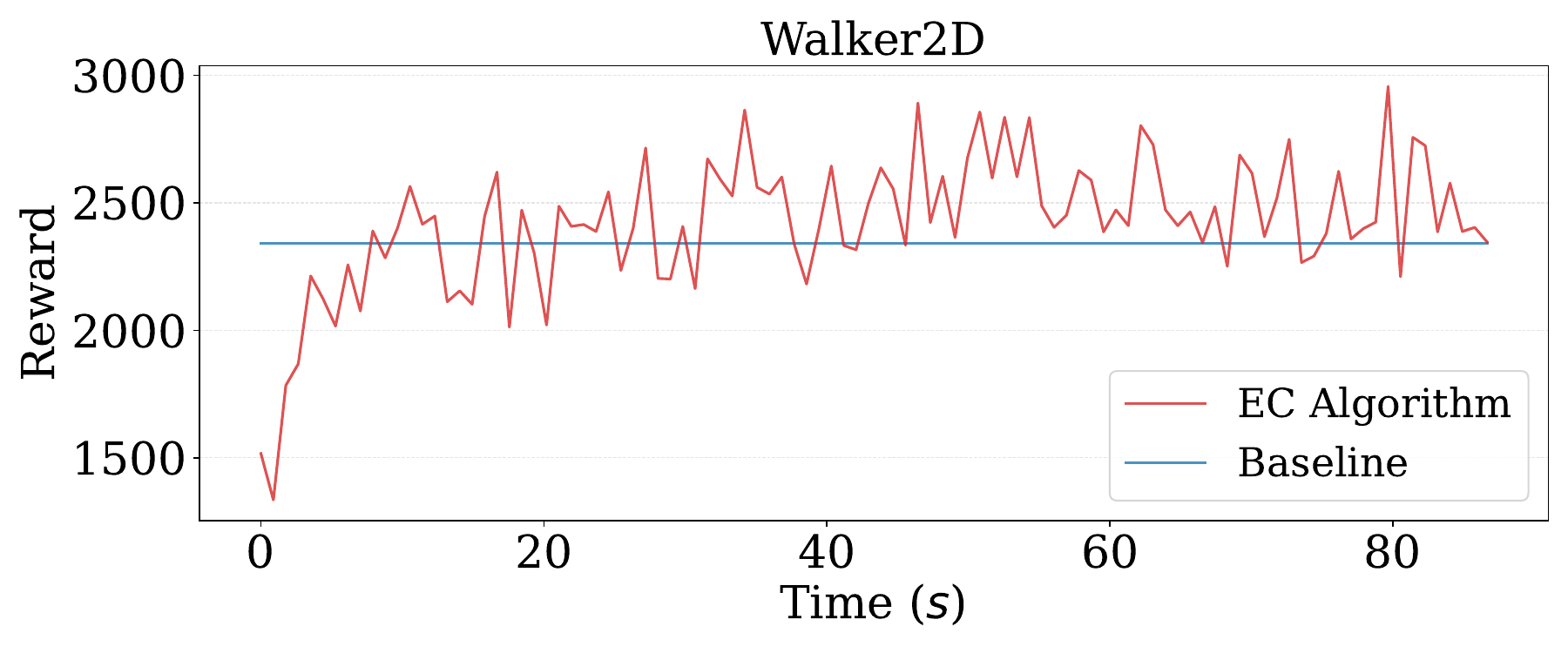}
    \caption{
    Performance curve when tackling the Hopper task with Brax. 
    The EC algorithm employed is CMA-ES with a population size of 4,096.
    The baseline performance, achieved by PPO2,  was directly taken from \cite{OpenAI_rl_baselines} as a reference.
    }
    \label{fig:rl_brax}
\end{figure}

Specifically, we instantiated a Walker2D task with the policy model being a 3-layer MLP with 1,830 parameters.
As shown in Fig.~\ref{fig:rl_brax}, with GPU acceleration, the EC algorithm was able to achieve the baseline performance within approximately 1 minute, underscoring the promising potential of \ourlib{} in tackling reinforcement learning tasks via neuroevolution.

\subsection{Comparison with EvoTorch}
\label{sec:exp_5}
To further benchmark the efficiency of \ourlib{}, we juxtaposed it against EvoTorch~\cite{toklu2023evotorch}, a library built atop PyTorch. 
We concentrated on evaluating two natively supported algorithms by both \ourlib{} and EvoTorch: PGPE and xNES. 

\begin{figure}
    \centering
    \begin{subfigure}{0.45\columnwidth}
        \centering
        \includegraphics[width=\columnwidth]{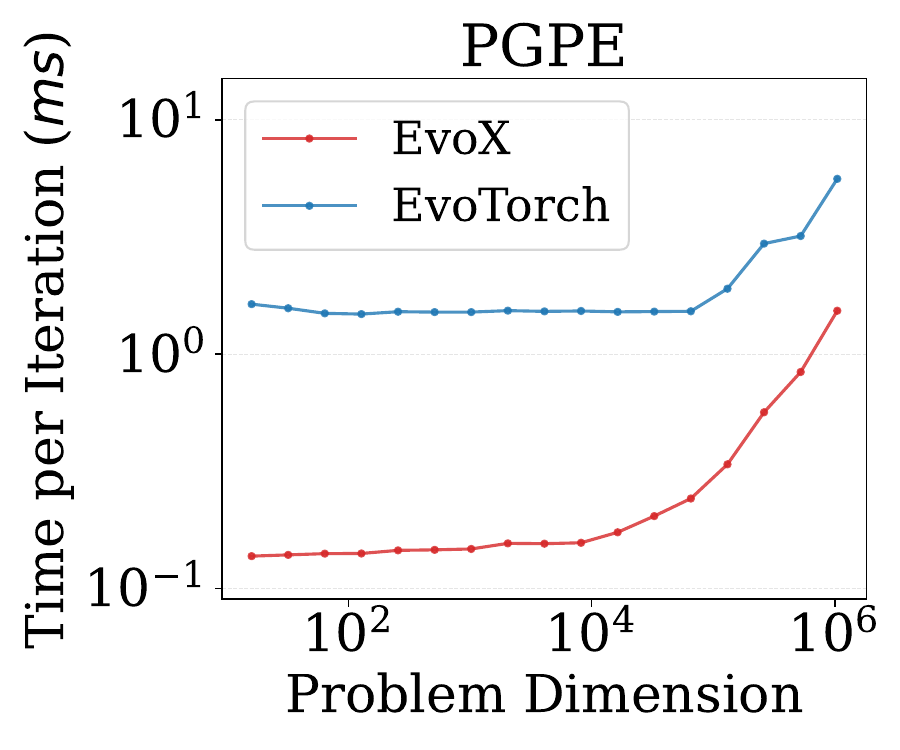}
    \end{subfigure}
    \begin{subfigure}{0.45\columnwidth}
        \centering
        \includegraphics[width=\columnwidth]{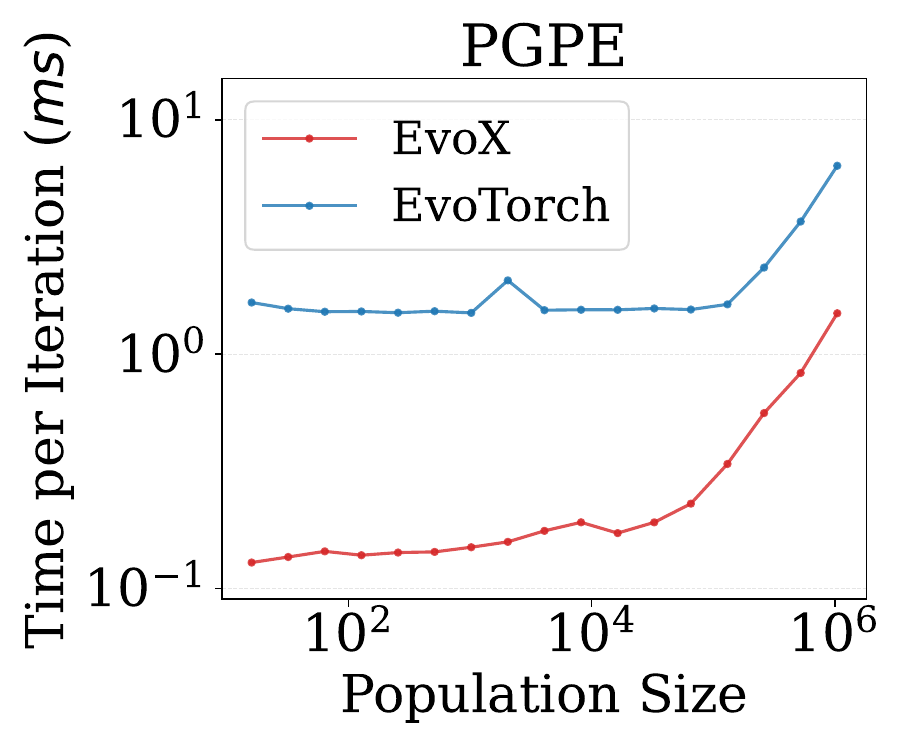}
    \end{subfigure}
    \begin{subfigure}{0.45\columnwidth}
        \centering
        \includegraphics[width=\columnwidth]{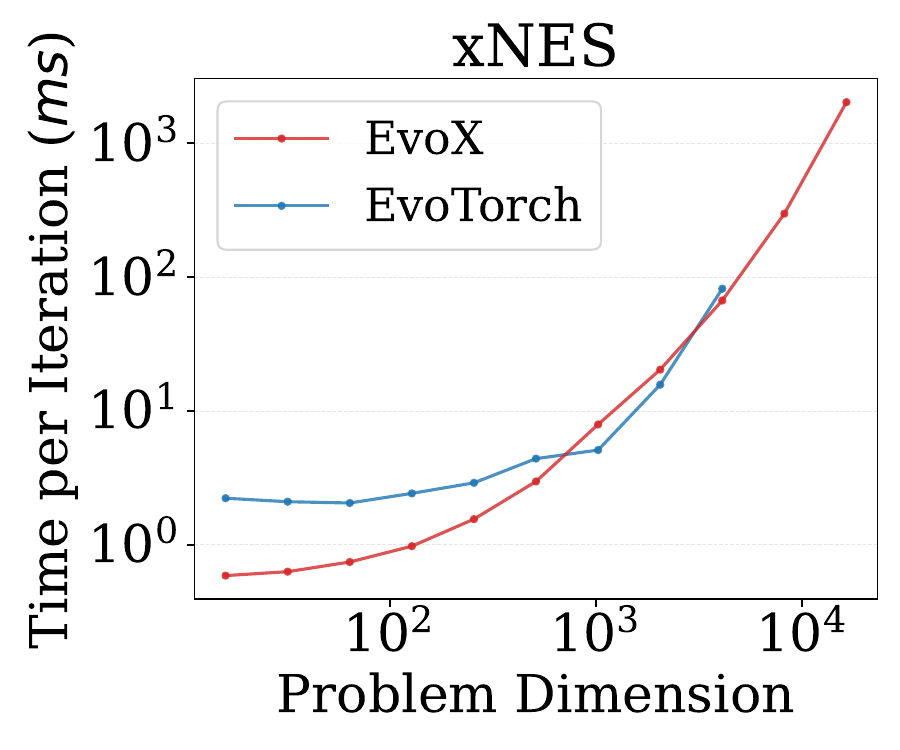}
    \end{subfigure}
    \begin{subfigure}{0.45\columnwidth}
        \centering
        \includegraphics[width=\columnwidth]{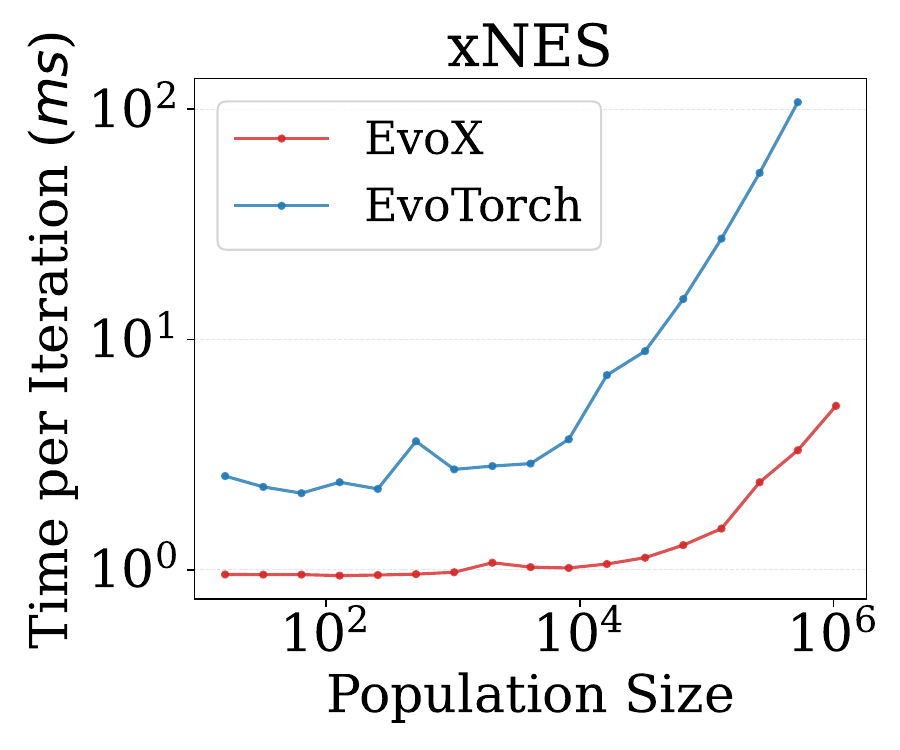}
    \end{subfigure}
    \caption{
    Comparative performance with EvoTorch using the Sphere function for PGPE and xNES. 
    Both axes are on a logarithmic scale. 
    For tests where the problem dimension is scaled, the population size is fixed at 100, and vice versa. 
    \textbf{Note}: EvoTorch's tests of xNES were terminated due to an \emph{Out of Memory} error.
    }

    \label{fig:vs_evotorch_xnes}
\end{figure}

As evidenced in Fig.~\ref{fig:vs_evotorch_xnes}, \ourlib{} exhibits promising performance in comparison to EvoTorch.
PGPE's execution on \ourlib{} is generally faster across various configurations.
In the case of xNES, \ourlib{} also tends to be more efficient, especially during population scaling tests.
When the dimension increased to 8,192, EvoTorch encountered an \emph{Out of Memory} error and thus failed to continue the test. 
A similar issue arose when the xNES population size was increased to 1,048,576. 
In contrast, \ourlib{} managed to complete the benchmark on the same hardware setup, indicating its efficiency in memory management.

\section{Conclusion}
\label{sec:conclusion}

Throughout its history, EC has proven to be an effective tool in addressing a wide array of problems across numerous domains. However, the rise of large-scale data and complex systems has posed significant scalability challenges for it. In response, we have developed \ourlib{}, a computing framework specifically crafted for scalable EC. 
This framework, with its tailored programming and computation models coupled with a hierarchical state management system, facilitates the efficient employment of distributed and heterogeneous computational resources.

Designed with flexibility and extensibility in mind, \ourlib{} is committed to ongoing development towards various application areas. 
Among the promising directions for future development are evolutionary multitasking~\cite{gupta_multifactorial_2016} and evolutionary transfer optimization~\cite{tan_evolutionary_2021}.
These areas, characterized by computationally intensive yet inherently parallelizable workloads, can significantly benefit from GPU acceleration. 
Furthermore, as advancements in computing architectures continue to reshape the technological landscape, \ourlib{} is poised for ongoing refinement. 
This will guarantee the sustained relevance of EC within the rapidly changing domain of AI.

\section*{Acknowledgement}
We thank Minyang Chen, Jiachun Li, Zhenyu Liang, Kebin Sun, Lishuang Wang, Haoming Zhang, and Mengfei Zhao for their efforts in helping with the implementations and testings. In particular, we thank Yansong Huang for his unique contributions to addressing the visualization issues related to the entire project.

\ifCLASSOPTIONcaptionsoff
  \newpage
\fi

\newpage
\bibliography{egbib} 
\bibliographystyle{IEEEtran}

\end{document}


\title{Neural Architecture Search as Multiobjective Optimization Benchmarks: Problem Formulation and Performance Assessment\\(Supplementary Materials)}

\author{Zhichao Lu,~\IEEEmembership{Member,~IEEE},
        Ran Cheng,~\IEEEmembership{Senior Member,~IEEE},
        Yaochu Jin,~\IEEEmembership{Fellow,~IEEE},\\
        Kay Chen Tan,~\IEEEmembership{Fellow,~IEEE},
        and~Kalyanmoy Deb,~\IEEEmembership{Fellow,~IEEE}
}

\markboth{IEEE Transactions on Evolutionary Computation,~Vol.~X, No.~X, July~2022}%
{Lu \MakeLowercase{\textit{et al.}}: A Sample Article Using IEEEtran.cls for IEEE Journals}


\maketitle

\input{8-appendix}

\ifCLASSOPTIONcaptionsoff
  \newpage
\fi

\bibliography{egbib} 
\bibliographystyle{IEEEtran}